\documentclass[10pt,twocolumn,letterpaper]{article}

\usepackage{cvpr}              
\definecolor{cvprblue}{rgb}{0.21,0.49,0.74}
\usepackage[pagebackref,breaklinks,colorlinks,allcolors=cvprblue]{hyperref}
\usepackage{multirow}
\usepackage{bbding}
\usepackage{pifont}
\usepackage{wasysym}
\usepackage{amssymb}
\usepackage{booktabs}
\usepackage{xcolor}
\usepackage{colortbl}
\usepackage{tabularx}
\usepackage{subcaption}
\usepackage{graphicx}
\usepackage[accsupp]{axessibility}  

\def\paperID{00007} 
\def\confName{CVPR}
\def\confYear{2026}

\title{TAP into the Patch Tokens: \\ Leveraging Vision Foundation Model Features for AI-Generated Image Detection}

\author{
Ahmed Abdullah$^{1}$ \qquad
Nikolas Ebert$^{1}$ \qquad
Oliver Wasenm\"uller$^{1}$ 
\\
$^{1}$Mannheim University of Applied Sciences, Germany\\
{\tt\small \{a.abdullah, n.ebert, o.wasenmueller\}@th-mannheim.de}}

\begin{document}
\maketitle
\begin{abstract}
Recent methods demonstrate that large-scale pretrained models, such as CLIP vision transformers, effectively detect AI-generated images (AIGIs) from unseen generative models when used as feature extractors.
Many state-of-the-art methods for AI-generated image detection build upon the original CLIP-ViT to enhance this generalization. Since CLIP's release, numerous vision foundation models (VFMs) have emerged, incorporating architectural improvements and different training paradigms. Despite these advances, their potential for AIGI detection and AI image forensics remains largely unexplored.
In this work, we present a comprehensive benchmark across multiple VFM families, covering diverse pretraining objectives, input resolutions, and model scales. We systematically evaluate their out-of-the-box performance for detecting fully-generated AI-images and AI-inpainted images, and discover that the best model outperforms the original CLIP by more than 12\% in accuracy, beating established approaches in the process. To fully leverage the features of a modern VFM, we propose a simple redesign of the classifier head by utilizing tunable attention pooling (TAP), which aggregates output tokens into a refined global representation. Integrating TAP with the latest VFMs yields substantial performance gains across several AIGI detection benchmarks, establishing a new state-of-the-art on two challenging benchmarks for in-the-wild detection of AI-generated and -inpainted images.
\end{abstract}

\section{Introduction}
\label{sec:intro}

\begin{figure}[t]
\centering
\includegraphics[width=\linewidth]{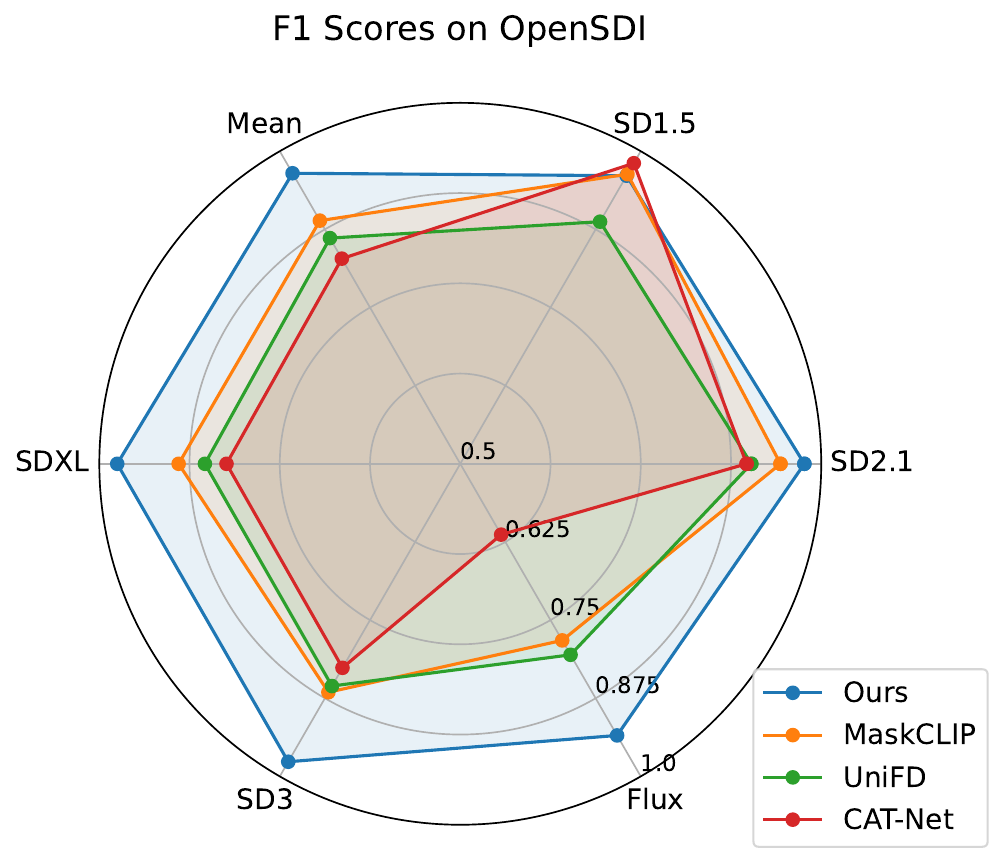}
\caption{Performance of our approach in terms of individual and mean F1 scores on the OpenSDI benchmark for AI-generated and -inpainted images \cite{wang2025opensdi}. Training only on SD1.5 \cite{rombach2022ldm}, we achieve state-of-the-art performance and generalization on unseen generator models (SD2.1 \cite{rombach2022ldm}, SD3 \cite{esser2024scalingsd3}, SDXL \cite{podell2023sdxl}, Flux \cite{flux2024}), surpassing established methods using only a frozen VFM \cite{bolya2025perception} and our tunable attention pooling (TAP).}
\label{fig:perf_pentagon}
\end{figure}

\begin{figure*}[t]
\centering

\begin{subfigure}{0.33\textwidth}
    \centering
    \includegraphics[width=\linewidth]{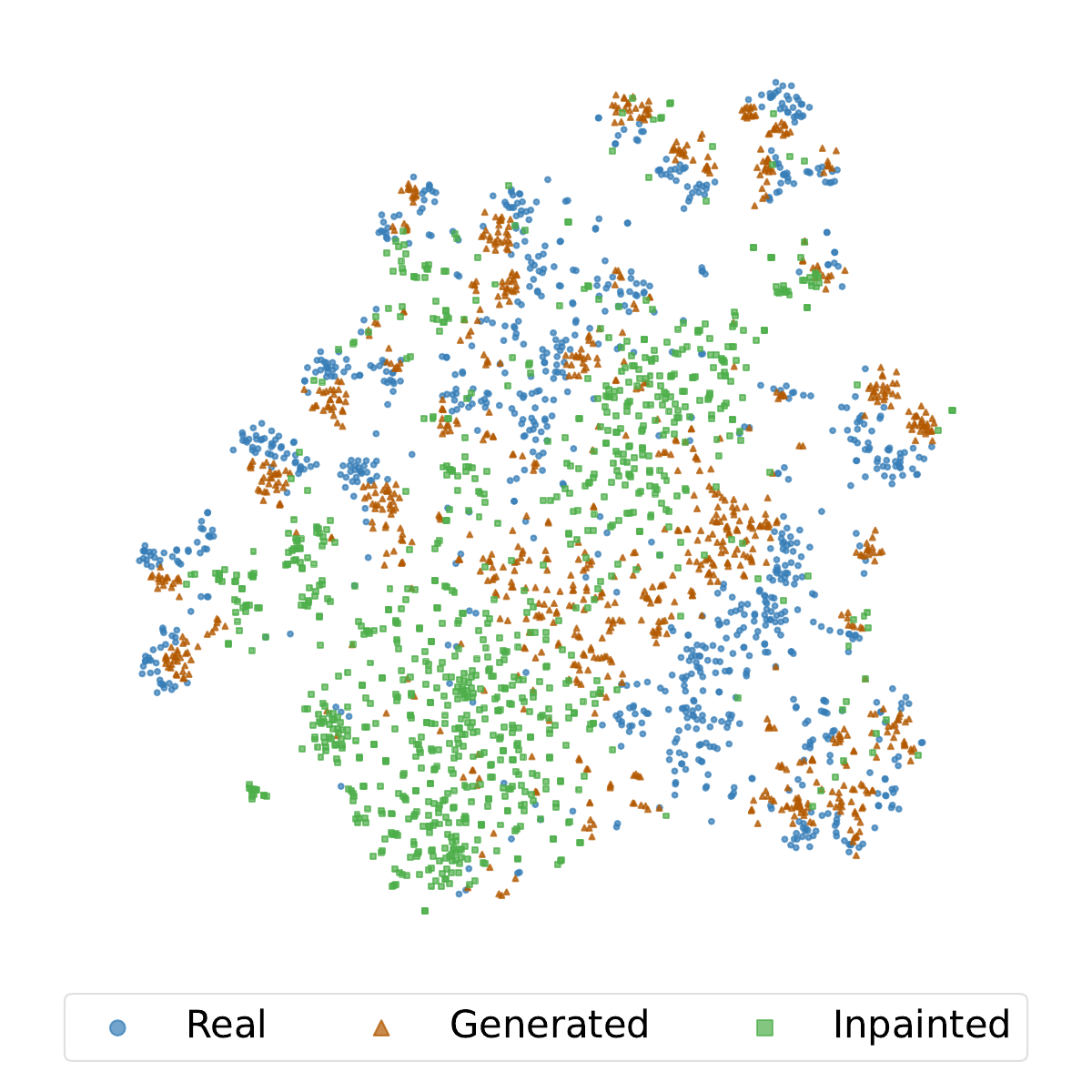}
    
    \label{fig:a}
\end{subfigure}
\hfill
\begin{subfigure}{0.33\textwidth}
    \centering
    \includegraphics[width=\linewidth]{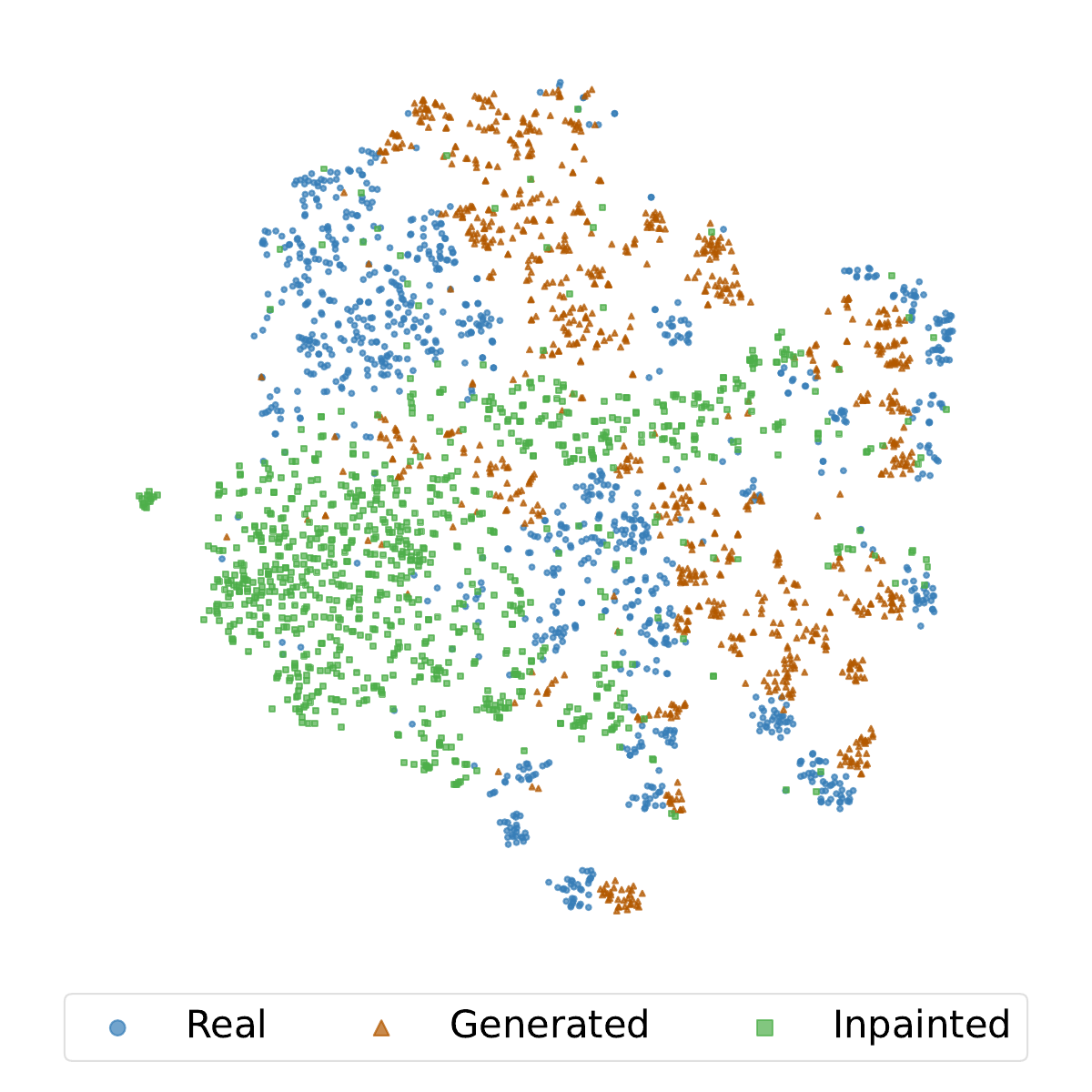}
    
    \label{fig:b}
\end{subfigure}
\hfill
\begin{subfigure}{0.33\textwidth}
    \centering
    \includegraphics[width=\linewidth]{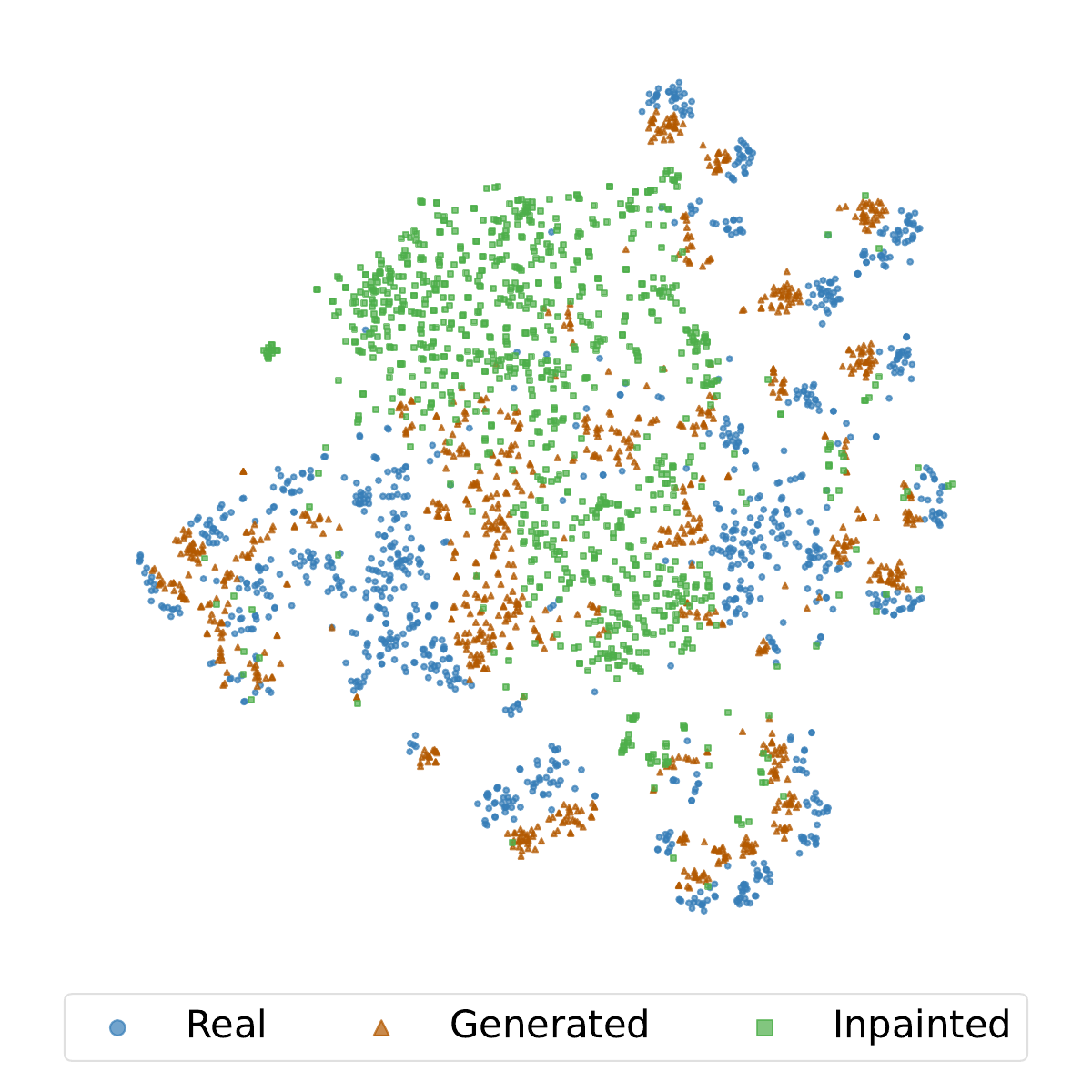}

    \label{fig:c}
\end{subfigure}
\hfill
\begin{subfigure}{0.33\textwidth}
    \centering
    \includegraphics[width=\linewidth]{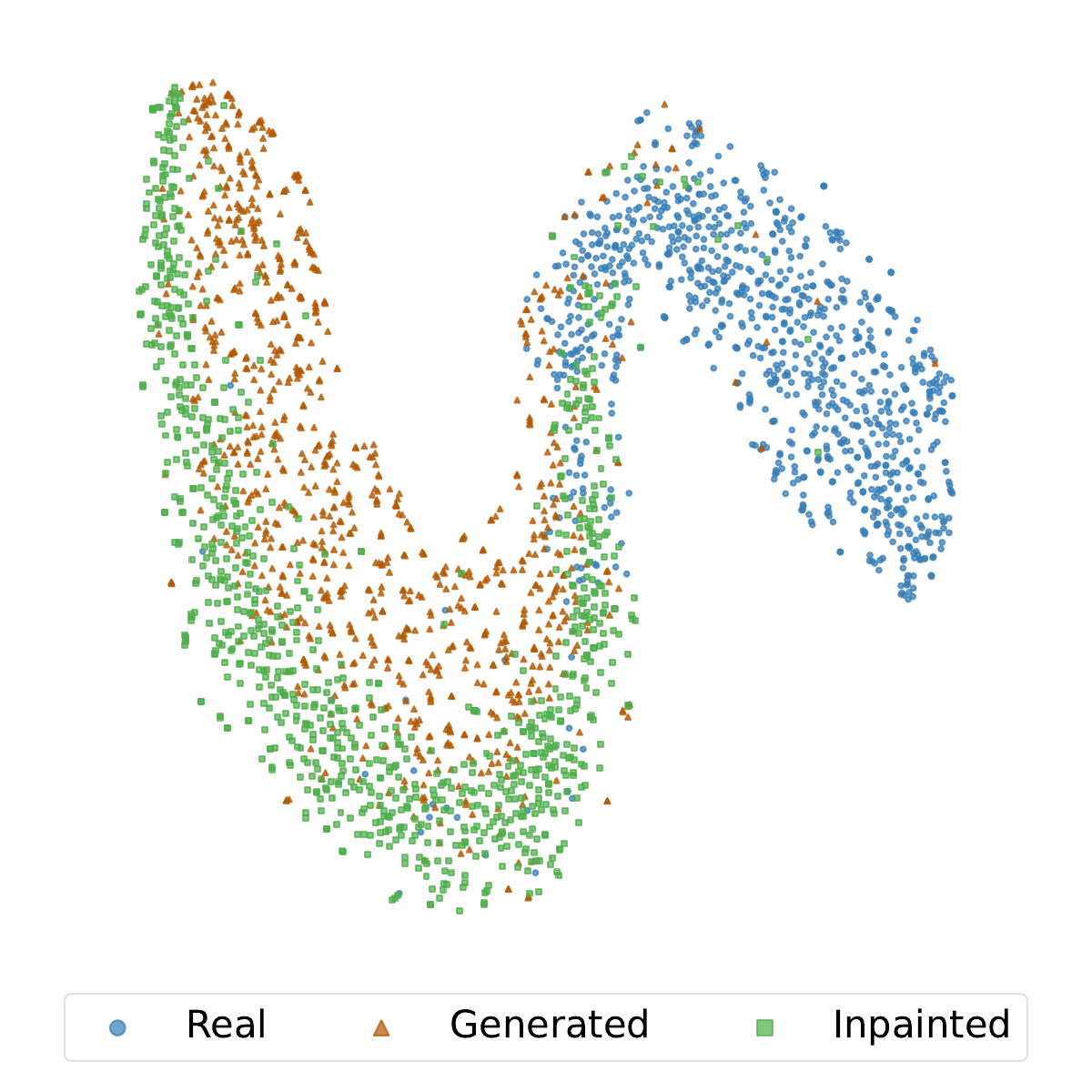}
    \caption{CLIP-ViT-L/14 \cite{radford2021clip}}
    \label{fig:d}
\end{subfigure}
\hfill
\begin{subfigure}{0.33\textwidth}
    \centering
    \includegraphics[width=\linewidth]{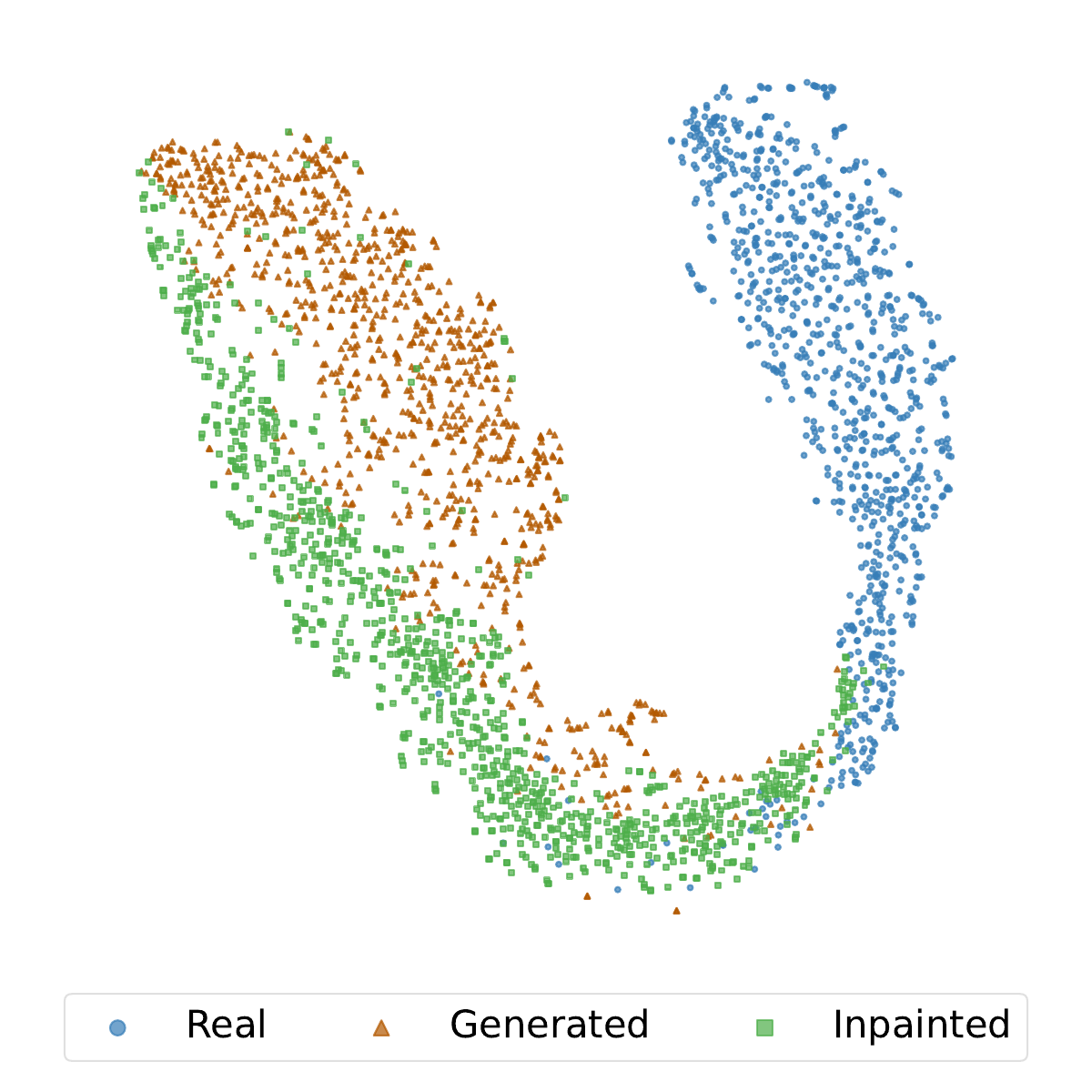}
    \caption{SIGLIP2-ViT-SO400M-NaFlex/16 \cite{siglip2}}
    \label{fig:e}
\end{subfigure}
\hfill
\begin{subfigure}{0.33\textwidth}
    \centering
    \includegraphics[width=\linewidth]{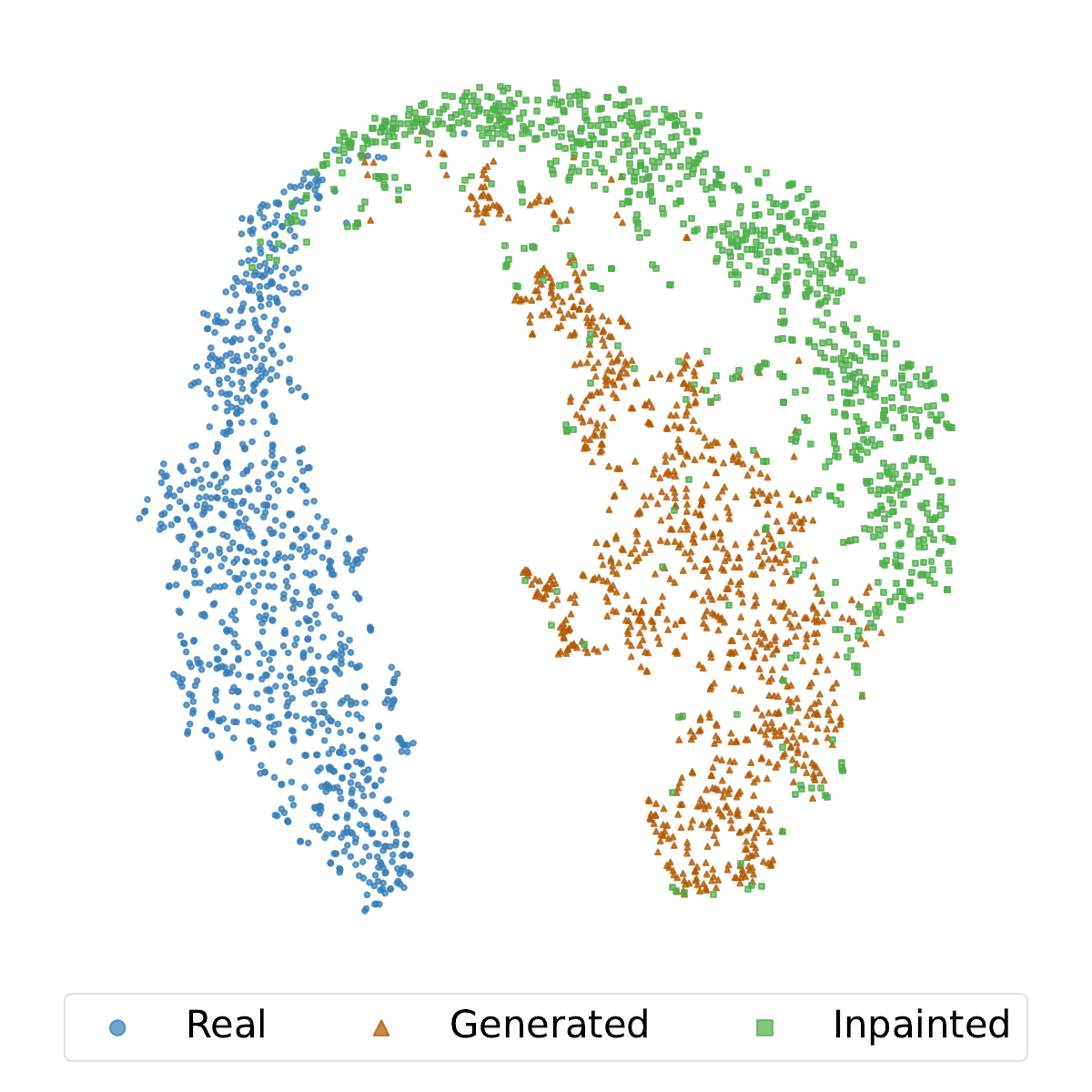}
    \caption{PE-Core-ViT-G/14 \cite{bolya2025perception}}
    \label{fig:f}
\end{subfigure}
\caption{t-SNE analysis of the feature spaces of three pretrained foundational ViT variants: CLIP-ViT-L/14 \cite{radford2021clip}, SIGLIP2-ViT-SO400M-NaFlex/16 \cite{siglip2}, and PE-Core-ViT-G/14 \cite{bolya2025perception}. Top row shows the $cls$ token output of the frozen visual encoders, while bottom row shows the global-pooled $gpl$ token obtained with our tunable attention pooling (TAP) approach. Features are obtained from the Stable~Diffusion~3 test set of OpenSDI \cite{wang2025opensdi}}
\label{fig:tsne}
\end{figure*}

Recently, the landscape of both proprietary and open-source AI image generation models has expanded rapidly. The latest generator models \cite{rombach2022ldm, flux2024, zhang2023addingcontrolnet, flux-2} continue to push the boundaries of photo-realistic image synthesis. 
However, this growth in capabilities has enabled malicious actors to misuse AI image generators in a wide range of attacks. For instance, they can be used to spread disinformation on social media, posing a serious threat to both the authenticity of online visual content and individual privacy. In addition to full-image synthesis, AI image generators today are capable of altering the content of a specific region within an input image (i.e. inpainting) with a single text prompt. This capability considerably extends the possible attack vectors for disinformation and fraud beyond full-image synthesis, as only a small fraction of the image may contain generative artifacts or semantic inconsistencies.

To counter this growing threat,  a variety of methods and benchmarks for AI-generated image (AIGI) detection have been proposed \cite{wang2020cnn, ojha2023unifd, tan2024npr, zhu2023genimage, wang2025opensdi, yan2024sanity}. A key challenge that recent methods aim to address is the generalization of detectors to unseen generative models. To achieve this, methods are trained to construct a feature space that can maximize generalization while exposed to only a single generator at train-time. Some approaches analyze pixel-level information in the RGB space \cite{ojha2023unifd, wang2020cnn}, while others target on detecting low-level artifacts in alternative image representations \cite{frank2020leveragingfreq, tan2024freqnet, tan2024npr}.

Since their inception, vision transformers (ViTs) \cite{dosovitskiy2021vit} have become ubiquitous in computer vision. They are widely used as the backbone for large-scale vision foundation models (VFMs) owing to their ability to scale effectively with both data and model size. Leveraging large-scale pretrained backbones has become a prominent trend in AI-generated image (AIGI) detection, with backbones trained through DINO \cite{oquab2023dinov2}, Masked Autoencoding \cite{he2022masked}, and CLIP \cite{radford2021clip} being used for their rich semantic features \cite{wang2025opensdi, tan2025c2pclip, ojha2023unifd}. Among these, CLIP has emerged as the most adopted model for AIGI detection, due to its unique ability to align visual and textual concepts.
Recently, many CLIP-style VFMs have been proposed that leverage improvements to the ViT architecture \cite{su2024roformer, beyer2023flexivit, dehghani2023patch}. Combined with even larger and more diverse training recipes, the new generation of VFMs demonstrate improved zero-shot classification and transfer to dense tasks \cite{bolya2025perception, siglip2}, significantly outperforming the original CLIP model. Despite this progress, it remains unclear how much these newer VFMs improve performance in AIGI detection. To address this, we present a comprehensive evaluation of modern VFMs for detecting both fully-generated and -inpainted images. Our experiments span multiple model families, pretraining objectives, ViT variants, and input resolutions, systematically addressing this question.
In addition, various methods for AIGI detection which employ CLIP for semantic feature extraction, do so by utilizing only a fraction of the full output sequence \cite{ojha2023unifd, koutlis2024rine, yermakov2026deepfakeclip}, namely, the $cls$ token, while discarding the patch tokens that encode local visual information. This design choice is particularly limiting for AIGI detection, where generative artifacts and inpainted manipulations often appear in localized image regions rather than across the image as a global pattern. As a result, relying solely on the $cls$ token may suppress fine-grained cues that are crucial for detection.

To this end, we propose a simple yet effective approach for semantic feature extraction that aligns the full output sequence (patch tokens and the $cls$ token) more closely with the AIGI detection task using a single tunable attention pooling (TAP) layer. 
Our approach keeps trainable parameters at minimum, while at the same time yielding significant improvements in generalization. Combining TAP features with the latest VFMs yields state-of-the-art detection performance on unseen generators (Figure \ref{fig:perf_pentagon}).

In essence, our contributions are as follows: (i) We carry out a comprehensive benchmark for measuring transferability of out-of-the-box features of modern VFMs to the AIGI detection task, and discover that many VFMs outclass not only CLIP \cite{radford2021clip}, but also various established detectors \cite{wang2025opensdi, tan2024npr, koutlis2024rine}.

(ii) We propose an updated design for semantic feature extraction, utilizing both patch tokens and the $cls$ token via tunable attention pooling (TAP). This simple redesign aligns output features closer to the AIGI detection task by enabling the detector to capture local generative artifacts while preserving global semantic patterns. (iii) We establish a new state-of-the-art on two challenging AIGI detection datasets, achieving an improvement of over 29\% and 10\% in accuracy over prior methods on Chameleon \cite{yan2024sanity} and OpenSDI \cite{wang2025opensdi}, respectively.

\section{Related Works}
\label{sec:related_works}

\begin{figure}[t]
\centering
\includegraphics[width=\linewidth]{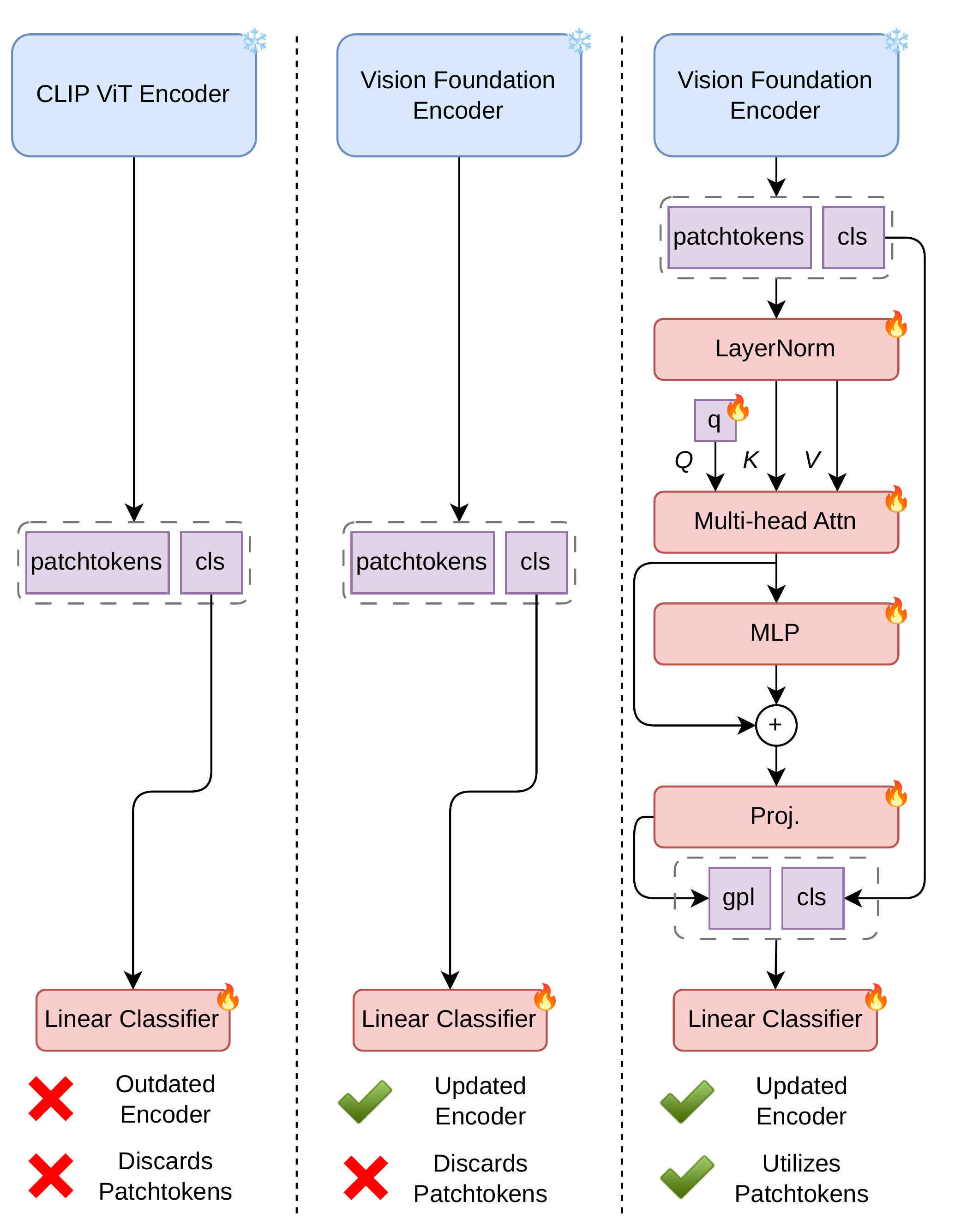}
\caption{Method Overview. Classical semantic feature extraction for AIGI detection relies on the outdated prior of the original CLIP ViT \cite{radford2021clip}. In addition, many approaches discard the patch tokens and utilize only the $cls$ token. Our approach updates the semantic feature extraction process by swapping the CLIP ViT with an encoder from a vision foundation model (VFM). Moreover, we take advantage of the full output sequence by employing tunable attention pooling (TAP), reducing the dimensionality while also obtaining a refined global embedding $gpl$, yielding improvements over a $cls$-token-only approach.}
\label{fig:method}
\end{figure}

\textbf{CLIP in AIGI Detection}.
Earlier methods for AIGI detection employed finetuned baseline CNN classifiers for detecting images synthesized by generative adversarial networks (GANs) \cite{wang2020cnn, zhang2019detecting, karras2019style}. As image generation methods evolved, and newer architectures such as latent diffusion models (LDMs) \cite{rombach2022ldm}, variational autoencoders (VAEs) \cite{van2017vqvae, razavi2019vqvae2}, and diffusion transformers (DiTs) \cite{peebles2023scalabledit, flux2024, flux-2} emerged, generalization to unseen architectures became even more essential. Pursuing a feature space that can achieve this quality, UniFD \cite{ojha2023unifd} proposed using the CLIP-pretrained vision transformer \cite{radford2021clip} for extracting semantic features for AIGI detection, and demonstrated improved generalization to LDMs and other generator models while being exposed only to images sourced from ProGAN \cite{karras2018progressiveprogan}. Following this finding, many works for AIGI detection started incorporating the CLIP-pretrained ViT in their methods \cite{Abdullah2026DualSight,tan2025c2pclip, smeu2025declip, koutlis2024rine, wang2025opensdi, yan2024sanity}. C2P-CLIP \cite{tan2025c2pclip} employs Low-Rank Adaptation (LoRA) \cite{zanella2024loravit, hulorallm} as a parameter-efficient fine-tuning strategy, together with distinct real and fake textual captions, to adapt the feature space of the CLIP ViT for AIGI detection while mitigating overfitting to the training data. 
DualSight \cite{Abdullah2026DualSight} extends this CLIP-based paradigm by combining a LoRA-tuned semantic encoder with a dedicated artifact encoder \cite{he2016resnet, ebert2023plg} operating on Sobel-filtered images, while progressively fusing representations from both encoders through cross-attention to jointly capture semantic inconsistencies and low-level generative artifacts. Similarly, AIDE \cite{yan2024sanity} adopts a dual-branch architecture in which average-pooled CLIP features provide a semantic prior, while artifact cues are extracted through a patch-wise frequency encoder, before both representations are concatenated for final prediction. 
In contrast, our approach focuses on an efficient redesign of the classifier head, without introducing additional encoders or fusion modules.

By utilizing a single tunable attention pooling layer (TAP), we are able to maximize the semantic information learned from the ViT by exploiting both the patch tokens and the $cls$ token, allowing the detector to emphasize spatial regions that may contain generative artifacts while maintaining global semantic information.

\textbf{Vision Foundation Model Landscape}.
Early visual representation learning relied on supervised pretraining on ImageNet \cite{deng2009imagenet}. Models pretrained on ImageNet learned transferable visual features that could be adapted to downstream tasks through fine-tuning, establishing this paradigm as the standard initialization strategy for many computer vision systems.
A significant shift occurred with multimodal pretraining approaches such as CLIP \cite{radford2021clip}, which jointly trains image and text encoders using large-scale image-caption pairs. By aligning visual and textual representations through contrastive learning, CLIP produces semantically rich embeddings that enable strong zero-shot generalization across diverse visual tasks. In parallel, self-supervised methods such as DINOv2 and DINOv3 \cite{oquab2023dinov2, dinov3} demonstrated that high-quality visual representations can be learned without labeled data. Using a teacher-student self-distillation framework on vision transformers \cite{dosovitskiy2021vit}, DINO captures meaningful semantic structures and object-level representations from large collections of unlabeled images.
Over the years, many VFM families were released, incorporating architectural improvements to the ViT such as rotary embedding \cite{su2024roformer}, variable input resolutions \cite{beyer2023flexivit, dehghani2023patch}, along with an even larger training scale \cite{siglip2, bolya2025perception, sam2}. However, it remains unclear how well the newest models transfer to the AIGI detection task, and whether or not a replacement of the original CLIP is overdue. While this question may have been partially addressed by the recent work of Zhou et al. \cite{zhou2025broughtgun}, their findings do not cover factors such as model scale, input resolution, performance on inpainted images, and utilization of features beyond the $cls$ token. Our comprehensive experiments offer a broader scope, and demonstrate the added benefit of integrating patch tokens within the detector framework.

\section{Method}
\label{sec:method}

\begin{table*}[t]
\centering
\small
\setlength{\tabcolsep}{6pt}
\renewcommand{\arraystretch}{1.1}

\begin{tabularx}{\linewidth}{Xccccccccc}
\toprule
\textbf{Method}
& \textbf{Midjourney} 
& \textbf{SD v1.4} 
& \textbf{SD v1.5} 
& \textbf{ADM} 
& \textbf{GLIDE} 
& \textbf{Wukong} 
& \textbf{VQDM} 
& \textbf{BigGAN} 
& \textbf{Mean} \\
\midrule

ResNet-50$^{\dagger}$ \cite{he2016resnet}
& 54.90 & 99.90 & 99.70 & 53.50 & 61.90 & 98.20 & 56.60 & 52.00 & 72.09 \\

Xception$^{\ddagger}$ \cite{chollet2017xception} 
& 57.97 & 98.06 & 97.98 & 51.16 & 57.51 & 97.79 & 50.34 & 48.74 & 69.94 \\

DeiT-S$^{\dagger}$ \cite{touvron2021deit} 
& 55.60 & 99.90 & 99.80 & 49.80 & 58.10 & 98.90 & 56.90 & 53.50 & 71.56 \\

Swin-T$^{\dagger}$ \cite{liu2021swin}
& 62.10 & 99.90 & 99.80 & 49.80 & 67.60 & 99.10 & 62.30 & 57.60 & 74.78 \\

CNNSpot$^{\dagger}$ \cite{wang2020cnn}
& 52.80 & 96.30 & 95.90 & 50.10 & 39.80 & 78.60 & 53.40 & 46.80 & 64.21 \\

Spec$^{\dagger}$ \cite{zhang2019detectingspec} 
& 52.00 & 99.40 & 99.20 & 49.70 & 49.80 & 94.80 & 55.60 & 49.80 & 68.79 \\

F3Net$^{\dagger}$ \cite{qian2020thinkingf3net}
& 50.10 & 99.90 & \textbf{99.90} & 49.90 & 50.00 & \underline{99.90} & 49.90 & 49.90 & 68.69 \\

GramNet$^{\dagger}$ \cite{liu2020gramnet} 
& 54.20 & 99.20 & 99.10 & 50.30 & 54.60 & 98.90 & 50.80 & 51.70 & 69.85 \\

DIRE$^{\dagger}$ \cite{wang2023dire} 
& 60.20 & 99.90 & 99.80 & 50.90 & 55.00 & 99.20 & 50.10 & 50.20 & 70.66 \\

UniFD$^{\dagger}$ \cite{ojha2023unifd} 
& 73.20 & 84.20 & 84.00 & 55.20 & 76.90 & 75.60 & 56.90 & 80.30 & 73.29 \\

GenDet$^{\dagger}$ \cite{zhu2023gendet}
& 89.60 & 96.10 & 96.10 & 58.00 & 78.40 & 92.80 & 66.50 & 75.00 & 81.56 \\

PatchCraft$^{\dagger}$ \cite{zhong2023patchcraft}
& 79.00 & 89.50 & 89.30 & 77.30 & 78.40 & 89.30 & 83.70 & 72.40 & 82.30 \\

NPR$^{\ddagger}$ \cite{tan2024npr} 
& 62.00 & 99.75 & 99.64 & 56.79 & 82.69 & 97.89 & 54.43 & 52.26 & 75.68 \\

SPSL$^{\ddagger}$ \cite{liu2021spatial} 
& 56.20 & 99.50 & 99.50 & 51.00 & 67.70 & 98.40 & 49.80 & 63.70 & 73.23 \\

SRM$^{\ddagger}$ \cite{luo2021generalizingsrm} 
& 54.10 & 99.80 & 99.80 & 49.90 & 52.80 & 99.60 & 50.00 & 51.00 & 69.63 \\

OMAT$^{\ddagger}$ \cite{zhou2025breakingomat}
& \underline{90.36} & 97.52 & 97.46 & 83.82 & \underline{97.41} & 97.62 & 95.53 & 97.34 & 94.63 \\

AIDE$^{\dagger}$ \cite{yan2024sanity} 
& 79.38 & 99.74 & 99.76 & 78.54 & 91.82 & 98.65 & 80.26 & 66.89 & 86.88 \\

\midrule
\textit{Ours} \\

PE-Core-ViT-L \cite{bolya2025perception}
& 82.42 & 99.72 & 99.53 & 65.97 & 84.69 & 99.05 & 92.97 & 92.74 & 89.64 \\

PE-Core-ViT-L \cite{bolya2025perception} + TAP
& 77.86 & \textbf{99.98} & \textbf{99.90} & 59.53 & 82.57 & 99.51 & 94.16 & 77.77 & 86.41 \\

PE-Core-ViT-G \cite{bolya2025perception}
& \textbf{94.92} & 99.92 & \underline{99.85} & \textbf{87.48} & 97.34 & 99.84 & \textbf{99.27} & \underline{98.68} & \textbf{97.16} \\

PE-Core-ViT-G \cite{bolya2025perception} + TAP
& 82.11 & \underline{99.96} & \textbf{99.90} & \underline{87.23} & \textbf{98.09} & \textbf{99.93} & \underline{99.19} & \textbf{98.94} & \underline{95.67} \\

\bottomrule
\end{tabularx}
\caption{Performance comparison on the GenImage \cite{zhu2023genimage} dataset. We benchmark the ViT-L and ViT-G variants of the PE-Core vision foundation model \cite{bolya2025perception} and compare accuracy with and without tunable attention pooling (TAP) features. All methods were trained on generated images sourced from SD v1.4 \cite{rombach2022ldm} with real images from ImageNet \cite{deng2009imagenet}. Results with '$^{\dagger}$' and '$^{\ddagger}$' are sourced from AIDE \cite{yan2024sanity} and OMAT \cite{zhou2025breakingomat}, respectively. Best results are in \textbf{bold}, while second best are \underline{underlined}.}
\label{tab:main_results_genimage}
\end{table*}

Our approach consists of two simple stages (Figure \ref{fig:method}). First, we extract semantic features using a frozen VFM and obtain a feature set containing the $cls$ token and the patch tokens. Next, we employ a classifier head with tunable attention pooling (TAP) for refining the initial features into a global-pooled $gpl$ token, which is used for classification alongside the $cls$ token.

\subsection{Semantic Feature Extraction}
Given input image $x$, we utilize a pretrained frozen vision encoder $\mathbf{Enc}$ from a VFM to extract a feature set $\mathbf{F}$ which consists of $N-1$ patch tokens, along with the $cls$ token. We denote the resulting feature set as

\begin{equation}
\mathbf{F} = \{\mathbf{f}_{cls}, \mathbf{f}_1, \dots, \mathbf{f}_{N-1}\} = \text{Enc}(x) \in \mathbb{R}^{N \times D},
\end{equation}
where $N$ is the total number of tokens, and $D$ is the embedding dimension of the VFM encoder. In Section \ref{ablations:vfm_benchmark}, we benchmark encoders from various VFM families to measure their out-of-the-box performance on AIGI detection using a linear classifier layer, and determine the best encoder for the task.

\subsection{Tunable Attention Pooling (TAP)}
To aggregate the feature set $\mathbf{F}$ into a compact representation, we employ a learnable pooling mechanism inspired from multihead attention pooling (MAP) \cite{zhai2022scaling}. Instead of relying on just the $cls$ token or simple average pooling, the model learns a query probe vector that selectively attends to relevant spatial features. A normalization layer is applied to stabilize the feature distribution before pooling.

The learnable probe vector $\mathbf{q} \in \mathbb{R}^{1 \times D}$ (randomly initialized) is used as the query in a multi-head cross-attention operation \cite{vaswani2017attention} over the feature set $\mathbf{F}$:

\begin{equation}
\mathbf{z} = \text{MHA}(\mathbf{q}, \text{LN}(\mathbf{F})),
\end{equation}
where $\text{LN}(\cdot)$ denotes layer normalization, and the normalized $\mathbf{F}$ provides the keys and values. This operation produces a pooled representation $\mathbf{z} \in \mathbb{R}^{1 \times D}$ that summarizes the most relevant spatial features. To further refine the representation, the pooled vector is passed through a two-layer MLP with a residual connection:

\begin{equation}
\mathbf{z}' = \mathbf{z} + \text{MLP}(\text{LN}(\mathbf{z}))
\end{equation}
Finally, the refined vector $\mathbf{z}'$ is passed through a linear projector to obtain the global-pooled token $gpl$. The $cls$ token is then concatenated with the resulting $gpl$ token and forwarded to the final linear layer for binary classification. The feature space analysis shown in Figure \ref{fig:tsne} compares the global-pooled $gpl$ token obtained using TAP against the $cls$ token from 3 different frozen encoders, showcasing enhanced feature separation between real, inpainted, and generated images. In Section \ref{TAP_impact}, we evaluate the TAP approach on multiple pretrained ViTs to measure the scalability of the pooling approach with model resolutions and number of patch tokens.

\section{Evaluations}
\label{sec:exps}
\begin{table*}[t]
\centering
\resizebox{\textwidth}{!}{%
\begin{tabular}{lcccccccccccc}
\toprule
 \multirow{2}{*}{\textbf{Method}} & \multicolumn{2}{c}{\textbf{SD1.5}} & \multicolumn{2}{c}{\textbf{SD2.1}} & \multicolumn{2}{c}{\textbf{SDXL}} & \multicolumn{2}{c}{\textbf{SD3}} & \multicolumn{2}{c}{\textbf{Flux}} & \multicolumn{2}{c}{\textbf{Mean}} \\ \cmidrule(lr){2-3} \cmidrule(lr){4-5} \cmidrule(lr){6-7} \cmidrule(lr){8-9} \cmidrule(lr){10-11} \cmidrule(lr){12-13}
 & F1 & Acc & F1 & Acc & F1 & Acc & F1 & Acc & F1 & Acc & F1 & Acc \\
\midrule
CNNSpot \cite{wang2020cnn}                & 84.60 & 85.04 & 71.56 & 75.94 & 59.70 & 68.72 & 56.27 & 67.08 & 35.72 & 57.57 & 61.57 & 70.87 \\
GramNet \cite{liu2020gramnet}            & 80.51 & 80.35 & 74.01 & 76.66 & 65.28 & 70.76 & 64.35 & 70.29 & 52.00 & 63.37 & 67.23 & 72.29 \\
FreqNet \cite{tan2024freqnet}            & 75.88 & 77.70 & 60.97 & 68.37 & 53.15 & 64.02 & 53.50 & 64.37 & 38.47 & 57.08 & 56.39 & 66.31 \\
NPR \cite{tan2024npr}                    & 79.41 & 79.28 & 81.67 & 81.84 & 72.12 & 74.28 & 73.43 & 75.47 & 67.62 & 71.36 & 74.85 & 76.45 \\
UniFD \cite{ojha2023unifd}               & 77.45 & 77.60 & 80.62 & 81.92 & 70.74 & 74.83 & 71.09 & 75.17 & 61.10 & 69.06 & 72.20 & 75.72 \\
RINE \cite{koutlis2024rine}              & 91.08 & 90.98 & 87.47 & 88.12 & 73.43 & 78.76 & 72.05 & 76.78 & 55.86 & 67.02 & 75.98 & 80.33 \\
MVSS-Net \cite{chen2021mvssnet}          & 93.47 & 93.65 & 79.27 & 82.33 & 59.85 & 70.42 & 62.80 & 72.13 & 27.59 & 56.78 & 64.60 & 75.06 \\
CAT-Net \cite{kwon2022catnet}            & \textbf{96.15} & \underline{96.15} & 79.32 & 82.46 & 64.76 & 73.34 & 65.26 & 73.61 & 22.66 & 55.26 & 65.63 & 76.16 \\
PSCC-Net \cite{liu2022psccnet}           & \underline{96.07} & 96.14 & 76.85 & 80.94 & 55.70 & 68.81 & 59.78 & 70.89 & 51.77 & 67.04 & 68.03 & 76.76 \\
ObjectFormer \cite{wang2022objectformer} & 71.72 & 75.22 & 66.79 & 72.55 & 49.19 & 62.92 & 48.32 & 62.54 & 37.92 & 58.05 & 54.79 & 66.26 \\
TruFor \cite{guillaro2023trufor}         & 90.12 & \textbf{97.73} & 35.93 & 55.62 & 58.04 & 66.41 & 59.73 & 67.51 & 49.12 & 61.62 & 58.59 & 69.78 \\
DeCLIP \cite{smeu2025declip}             & 80.68 & 78.31 & 84.02 & 82.77 & 70.69 & 70.55 & 69.93 & 68.40 & 51.77 & 65.61 & 71.42 & 73.13 \\
IML-ViT \cite{ma2023imlvit}              & 94.47 & 75.73 & 69.70 & 61.19 & 40.98 & 49.95 & 44.69 & 51.25 & 18.20 & 43.62 & 53.61 & 56.35 \\
MaskCLIP \cite{wang2025opensdi}          & 92.64 & 92.72 & 88.71 & 89.45 & 78.02 & 81.22 & 73.07 & 78.01 & 56.49 & 68.50 & 77.79 & 81.98 \\
DualSight \cite{Abdullah2026DualSight}                         & 85.50 & 84.96 & 93.41 & 93.53 & 92.02 & 92.25 & 92.64 & 92.86 & 79.30 & 82.06 & 88.57 & 89.13 \\

\midrule
\textit{Ours} \\

PE-Core-ViT-L \cite{bolya2025perception}
& 80.40 & 77.44 & 89.88 & 88.99 & 89.50 & 88.63 & 89.30 & 88.48 & \textbf{88.49} & \textbf{87.67} & 87.52 & 86.24 \\

PE-Core-ViT-L \cite{bolya2025perception} + TAP
& 89.19 & 88.80 & \underline{94.46} & \underline{94.42} & \underline{94.18} & \underline{94.14} & \underline{93.60} & \underline{93.60} & 86.78 & \underline{87.58} & \underline{91.64} & \underline{91.71} \\

PE-Core-ViT-G \cite{bolya2025perception}
& 87.11 & 86.68 & 91.86 & 92.06 & 91.83 & 92.05 & 90.48 & 90.87 & 86.43 & 87.40 & 89.54 & 89.81 \\

PE-Core-ViT-G \cite{bolya2025perception} + TAP
& 92.16 & 91.68 & \textbf{95.32} & \textbf{95.23} & \textbf{95.03} & \textbf{94.96} & \textbf{95.31} & \textbf{95.25} & \underline{86.89} & \textbf{87.67} & \textbf{92.94} & \textbf{92.96} \\

\bottomrule
\end{tabular}
}
\caption{Performance comparison on the OpenSDI \cite{wang2025opensdi} dataset for generated and inpainted images. We benchmark the ViT-L and ViT-G variants of the PE-Core vision foundation model \cite{bolya2025perception} and compare F1-score and accuracy with and without tunable attention pooling (TAP) features against prior methods. All methods are trained on the SD v1.5 training set of OpenSDI. Best results are in \textbf{bold}, while second best are \underline{underlined}.}
\label{tab:main_results_opensdi}
\end{table*}
\begin{table}[h]
\centering
\begin{tabularx}{\linewidth}{Xccc}
\hline
\textbf{Method} & \textbf{Acc} & \textbf{F-Acc} & \textbf{R-Acc }\\ \hline
CNNSpot    \cite{wang2020cnn}        & 60.11    & 8.86 & 98.63                 \\
GramNet    \cite{liu2020gramnet}     & 60.95    & 17.65 & 93.50                \\
LNP        \cite{liu2022detectinglnp}  & 55.63  & 0.57 & 97.01                 \\
UniFD     \cite{ojha2023unifd}        & 55.62  & 74.97 & 41.09                \\
DIRE       \cite{wang2023dire}         & 59.71  & 11.86 & 95.67                \\
PatchCraft \cite{zhong2023patchcraft}    & 56.32             & 3.07 & 96.35                 \\
NPR        \cite{tan2024npr}      & 58.13             & 2.43 & \textbf{100.00}                \\
AIDE       \cite{yan2024sanity}      & 62.60             & 20.33 & 94.38                \\
OMAT       \cite{zhou2025breakingomat}     & 66.05             & 33.93 & 90.17       \\ 
\midrule
\textit{Ours} \\

PE-ViT-L \cite{bolya2025perception}
& 72.81 & \underline{98.43} & 53.55 \\

PE-ViT-L \cite{bolya2025perception} + TAP
& 83.31 & 61.36 & \underline{99.82} \\

PE-ViT-G \cite{bolya2025perception}
& \underline{92.11} & \textbf{99.19} & 86.78 \\

PE-ViT-G \cite{bolya2025perception} + TAP
& \textbf{95.64} & 91.47 & 98.78 \\

\hline
\end{tabularx}
\caption{Performance Comparison on the Chameleon \cite{yan2024sanity} dataset. We benchmark the ViT-L and ViT-G variants of the PE-Core vision foundation model \cite{bolya2025perception} and compare accuracy (Acc), accuracy on generated images (F-Acc), and accuracy on real images (R-Acc) with and without tunable attention pooling (TAP) features. All methods are trained using the SD v1.4 training set from GenImage \cite{zhu2023genimage}. Best result is shown in \textbf{bold}, while second best is \underline{underlined}.}
\label{tab:main_results_chameleon}
\end{table}

\begin{table*}[t]
\centering
\scriptsize
\begin{tabularx}{\linewidth}{Xllcccccccccccc}
\toprule
 \multirow{2}{*}{\textbf{Model}} & \multirow{2}{*}{\textbf{Encoder}}& \multirow{2}{*}{\textbf{Res.}} & \multicolumn{2}{c}{\textbf{SD1.5}} & \multicolumn{2}{c}{\textbf{SD2.1}} & \multicolumn{2}{c}{\textbf{SDXL}} & \multicolumn{2}{c}{\textbf{SD3}} & \multicolumn{2}{c}{\textbf{Flux}} & \multicolumn{2}{c}{\textbf{Mean}} \\ \cmidrule(lr){4-5} \cmidrule(lr){6-7} \cmidrule(lr){8-9} \cmidrule(lr){10-11} \cmidrule(lr){12-13} \cmidrule(lr){14-15}
 
 & & & F1 & Acc & F1 & Acc & F1 & Acc & F1 & Acc & F1 & Acc & F1 & Acc \\
\midrule
\multirow{8}{*}{DINOv2 \cite{oquab2023dinov2}} & \multirow{4}{*}{ViT-L/14$^{\dagger}$} & 224 & 71.47 & 65.61 & 75.52 & 71.72 & 74.44 & 70.86 & 68.58 & 65.69 & 62.64 & 61.40 & 70.53 & 67.05 \\
& & 448 & 72.95 & 70.04 & 78.01 & 76.97 & 75.53 & 74.90 & 67.60 & 68.81 & 54.42 & 60.26 & 69.70 & 70.20 \\
& & 672 & 72.85 & 71.37 & 76.39 & 76.65 & 72.01 & 73.30 & 62.87 & 67.00 & 46.81 & 57.95 & 66.19 & 69.25 \\
& & 896 & 69.57 & 69.61 & 72.08 & 74.13 & 63.14 & 67.97 & 53.70 & 62.48 & 39.01 & 55.33 & 59.50 & 65.90 \\
\cmidrule(lr){2-15}
&\multirow{4}{*}{ViT-G/14$^{\dagger}$} & 224 & 73.64 & 69.76 & 78.47 & 76.56 & 77.43 & 75.81 & 72.33 & 71.54 & 64.12 & 65.35 & 73.20 & 71.80 \\
 && 448 & 76.18 & 74.36 & 81.73 & 81.50 & 79.91 & 80.05 & 70.42 & 72.72 & 54.95 & 63.12 & 72.64 & 74.35 \\
 && 672 & 77.08 & 76.44 & 80.29 & 81.15 & 77.36 & 79.02 & 64.94 & 70.44 & 44.36 & 59.45 & 68.81 & 73.30  \\
 && 896 & 75.87 & 75.96 & 78.77 & 80.22 & 72.51 & 75.90 & 58.58 & 67.14 & 38.14 & 57.20 & 64.77 & 71.28 \\
 \midrule
\multirow{16}{*}{DINOv3 \cite{dinov3}}&\multirow{4}{*}{ViT-L/16$^{\dagger}$} & 256 & 78.78 & 76.21 & 85.09 & 84.46 & 83.08 & 82.64 & 78.77 & 78.97 & 74.18 & 75.33 & 79.98 & 79.52 \\
 && 512 & 78.98 & 78.16 & 85.11 & 85.41 & 82.51 & 83.35 & 75.65 & 77.99 & 67.84 & 72.53 & 78.02 & 79.49 \\
 && 768 & 78.88 & 78.70 & 84.99 & 85.64 & 81.59 & 82.96 & 75.45 & 78.24 & 65.45 & 71.55 & 77.27 & 79.42 \\
& & 1024 & 79.55 & 79.54 & 85.09 & 85.70 & 81.58 & 82.88 & 72.85 & 76.46 & 61.19 & 68.99 & 76.05 & 78.71 \\
\cmidrule(lr){2-15}
&\multirow{4}{*}{ConvNext-Large$^{\dagger}$} & 256 & 76.26 & 72.34 & 81.82 & 79.95 & 80.60 & 78.80 & 78.62 & 77.07 & 75.35 & 74.26 & 78.53 & 76.48 \\
& & 512 & 76.84 & 74.55 & 83.36 & 82.65 & 81.62 & 81.21 & 78.10 & 78.30 & 71.41 & 73.06 & 78.27 & 77.95 \\
& & 768 & 77.37 & 76.50 & 83.76 & 84.18 & 81.79 & 82.66 & 76.69 & 78.63 & 63.38 & 69.56 & 76.60 & 78.31\\
& & 1024 & 79.43 & 78.84 & 85.90 & 86.31 & 82.71 & 83.70 & 72.19 & 75.90 & 53.88 & 64.98 & 74.82 & 77.95 \\
\cmidrule(lr){2-15}
&\multirow{4}{*}{ViT-H+/16$^{\dagger}$} & 256 & 77.87 & 76.51 & 83.80 & 83.90 & 81.72 & 82.09 & 76.89 & 78.34 & 72.04 & 74.66 & 78.46 & 79.10 \\
& & 512 & 77.72 & 77.66 & 83.86 & 84.54 & 80.92 & 82.20 & 75.89 & 78.38 & 69.42 & 73.90 & 77.56 & 79.34 \\
 && 768 & 78.27 & 76.81 & 84.22 & 84.03 & 82.16 & 82.31 & 79.34 & 79.99 & 73.39 & 75.38 & 79.48 & 79.70 \\
& & 1024 & 75.59 & 76.83 & 82.51 & 83.98 & 78.88 & 81.28 & 72.09 & 76.52 & 63.39 & 71.11 & 74.49 & 77.94 \\
\cmidrule(lr){2-15}
&\multirow{4}{*}{ViT-7B/16$^{\dagger}$} & 256 & 79.85 & 77.07 & 86.21 & 85.45 & 85.52 & 84.81 & 83.76 & 83.21 & 80.07 & 79.99 & 83.08 & 82.11 \\
& & 512 & 81.90 & 81.74 & 87.21 & 87.74 & 85.04 & 85.99 & 80.61 & 82.47 & 71.89 & 76.28 & 81.33 & 82.84 \\
& & 768 & 82.63 & 82.77 & 87.36 & 87.98 & 84.53 & 85.68 & 81.03 & 82.92 & 71.15 & 75.98 & 81.34 & 83.07 \\
& & 1024 & \underline{83.00} & \underline{83.72} & 87.29 & 88.17 & 83.98 & 85.57 & 79.63 & 82.26 & 67.01 & 74.00 & 80.18 & 82.75 \\
\midrule
SAM2.1 \cite{sam2} & Hiera-Large & 1024 & 79.97 & 79.74 & 78.21 & 79.11 & 60.56 & 66.67 & 54.65 & 63.56 & 38.12 & 55.39 & 62.30 & 68.89 \\
\midrule
\multirow{2}{*}{CLIP \cite{radford2021clip}}&\multirow{2}{*}{ViT-L/14} & 224 & 68.78 & 63.59 & 73.29 & 70.00 & 70.90 & 67.86 & 71.05 & 68.02 & 68.69 & 66.04 & 70.54 & 67.10 \\
              & & 336 & 70.72 & 71.05 & 79.99 & 81.37 & 78.63 & 80.33 & 77.03 & 79.14 & 71.77 & 75.37 & 75.63 & 77.45 \\
\midrule
\multirow{13}{*}{SIGLIP2 \cite{siglip2}}&\multirow{3}{*}{ViT-L/16} & 256 & 76.27 & 74.32 & 81.47 & 81.30 & 80.59 & 80.62 & 79.86 & 80.03 & 75.05 & 76.18 & 78.65 & 78.49 \\
& & 384 & 78.90 & 77.88 & 84.85 & 85.37 & 85.97 & 86.28 & 84.27 & 84.96 & 77.90 & 79.83 & 82.38 & 82.87 \\
& & 512 & 81.21 & 79.29 & 88.64 & 88.40 & 89.17 & 88.91 & 89.22 & 89.01 & \underline{86.63} & 86.64 & 86.97 & 86.45 \\
\cmidrule(lr){2-15}
&\multirow{3}{*}{ViT-SO400M/16} & 256 & 77.43 & 73.83 & 83.36 & 82.07 & 82.47 & 81.22 & 81.96 & 80.77 & 77.20 & 76.67 & 80.48 & 78.91 \\
& & 384 & 81.11 & 79.84 & 87.87 & 87.96 & 88.04 & 88.10 & 87.91 & 88.04 & 82.07 & 83.03 & 85.40 & 85.39 \\
& & 512 & 81.21 & 80.17 & 88.62 & 88.82 & 89.11 & \underline{89.28} & \underline{89.32} & \underline{89.48} & 85.08 & 85.83 & 86.67 & \underline{86.72} \\
\cmidrule(lr){2-15}
&\multirow{2}{*}{ViT-SO400M/14}&  224 & 76.76 & 72.70 & 82.61 & 80.99 & 82.29 & 80.67 & 81.18 & 79.69 & 77.58 & 76.37 & 80.08 & 78.08 \\
& & 378 & 80.10 & 79.23 & 86.79 & 87.27 & 87.14 & 87.52 & 85.82 & 86.50 & 79.40 & 81.35 & 83.85 & 84.37 \\
\cmidrule(lr){2-15}
&\multirow{3}{*}{ViT-SO400M-NaFlex/16$^{\dagger}$} & 256 & 74.68 & 71.78 & 82.62 & 81.79 & 82.17 & 81.38 & 81.63 & 80.81 & 77.86 & 77.62 & 79.79 & 78.68 \\
& & 384 & 75.24 & 75.14 & 83.66 & 84.43 & 83.36 & 84.20 & 82.86 & 83.82 & 74.74 & 77.58 & 79.97 & 81.03 \\
& & 512 & 76.58 & 76.90 & 83.54 & 84.88 & 83.74 & 85.02 & 83.79 & 85.11 & 77.30 & 80.24 & 80.99 & 82.43 \\
\cmidrule(lr){2-15}
&\multirow{2}{*}{ViT-gopt/16} & 256 & 78.01 & 74.20 & 83.22 & 81.56 & 82.65 & 81.04 & 81.78 & 80.17 & 78.15 & 77.01 & 80.76 & 78.80 \\
& & 384 & 80.51 & 79.88 & 86.70 & 87.20 & 86.92 & 87.30 & 86.76 & 87.19 & 80.53 & 82.17 & 84.28 & 84.75 \\
\midrule
\multirow{2}{*}{PE-Core \cite{bolya2025perception}}& ViT-L/14 & 336 & 80.40 & 77.44 & \underline{89.88} & \underline{88.99} & \underline{89.50} & 88.63 & 89.30 & 88.48 & \textbf{88.49} & \textbf{87.67} & \underline{87.52} & 86.24 \\
\cmidrule(lr){2-15}
& ViT-G/14  & 448 & \textbf{87.11} & \textbf{86.68} & \textbf{91.86} & \textbf{92.06} & \textbf{91.83} & \textbf{92.05} & \textbf{90.48} & \textbf{90.87} & 86.43 & \underline{87.40} & \textbf{89.54} & \textbf{89.81} \\
\bottomrule
\end{tabularx}
\caption{Comparison of visual encoders from various vision foundation model families on OpenSDI \cite{wang2025opensdi}. $^{\dagger}$ indicates a variable-resolution model. Best results are in \textbf{bold}, while second best are \underline{underlined}.}
\label{tab:visual_res}
\end{table*}

\begin{table*}[t]
\centering
\small
\setlength{\tabcolsep}{4pt}
\renewcommand{\arraystretch}{1.15}

\resizebox{\textwidth}{!}{
\begin{tabular}{llc c cc cc cc cc cc cc}
\toprule
\multirow{2}{*}{\textbf{ViT}} 
& \multirow{2}{*}{\textbf{Model}} 
& \multirow{2}{*}{\textbf{Res/Patch}} 
& \multirow{2}{*}{\textbf{TAP}} 
& \multicolumn{2}{c}{\textbf{SD1.5}} 
& \multicolumn{2}{c}{\textbf{SD2.1}} 
& \multicolumn{2}{c}{\textbf{SDXL}} 
& \multicolumn{2}{c}{\textbf{SD3}} 
& \multicolumn{2}{c}{\textbf{Flux}} 
& \multicolumn{2}{c}{\textbf{Mean}} \\
\cmidrule(lr){5-6}
\cmidrule(lr){7-8}
\cmidrule(lr){9-10}
\cmidrule(lr){11-12}
\cmidrule(lr){13-14}
\cmidrule(lr){15-16}
& & & 
& F1 & Acc
& F1 & Acc
& F1 & Acc
& F1 & Acc
& F1 & Acc
& F1 & Acc \\
\midrule

\multirow{6}{*}{L}

& \multirow{2}{*}{CLIP \cite{radford2021clip}}
& \multirow{2}{*}{336/14}
& \ding{55}
& 70.72 & 71.05 & 79.99 & 81.37 & 78.63 & 80.33 & 77.03 & 79.14 & 71.77 & 75.37 & 75.63 & 77.45 \\
& & & \ding{51}
& 75.35 & 76.46 & 87.63 & 88.08 & 86.47 & 87.05 & 85.76 & 86.56 & 75.08 & 78.49 & 82.06 & 83.33 \\

\cmidrule(lr){2-16}

& \multirow{2}{*}{SigLIP2 \cite{siglip2}}
& \multirow{2}{*}{512/16}
& \ding{55}
& 81.21 & 79.29 & 88.64 & 88.40 & 89.17 & 88.91 & 89.22 & 89.01 & 86.63 & 86.64 & 86.97 & 86.45 \\
& & & \ding{51}
& 85.90 & 85.22 & 93.27 & 93.23 & 92.23 & 92.22 & 93.14 & 93.13 & \underline{87.03} & \underline{87.65} & 90.31 & 90.29 \\

\cmidrule(lr){2-16}

& \multirow{2}{*}{PE \cite{bolya2025perception}}
& \multirow{2}{*}{336/14}
& \ding{55}
& 80.40 & 77.44 & 89.88 & 88.99 & 89.50 & 88.63 & 89.30 & 88.48 & \textbf{88.49} & \textbf{87.67} & 87.52 & 86.24 \\
& & & \ding{51}
& \underline{89.19} & \underline{88.80} & \underline{94.46} & \underline{94.42} & \underline{94.18} & \underline{94.14} & \underline{93.60} & \underline{93.60} & 86.78 & 87.58 & \underline{91.64} & \underline{91.71} \\

\midrule

\multirow{2}{*}{G}
& \multirow{2}{*}{PE \cite{bolya2025perception}}
& \multirow{2}{*}{448/14}
& \ding{55}
& 87.11 & 86.68 & 91.86 & 92.06 & 91.83 & 92.05 & 90.48 & 90.87 & 86.43 & 87.40 & 89.54 & 89.81 \\
& & & \ding{51}
& \textbf{92.16} & \textbf{91.68} & \textbf{95.32} & \textbf{95.23} & \textbf{95.03} & \textbf{94.96} & \textbf{95.31} & \textbf{95.25} & 86.89 & \textbf{87.67} & \textbf{92.94} & \textbf{92.96} \\

\midrule

\multirow{6}{*}{\begin{tabular}[c]{@{}l@{}}SO400M\\(NaFlex)\end{tabular}}
& \multirow{6}{*}{SigLIP2 \cite{siglip2}}

& \multirow{2}{*}{256/16}
& \ding{55}
& 74.68 & 71.78 & 82.62 & 81.79 & 82.17 & 81.38 & 81.63 & 80.81 & 77.86 & 77.62 & 79.79 & 78.68 \\
& & & \ding{51}
& 80.30 & 78.64 & 87.04 & 86.62 & 85.59 & 85.28 & 85.92 & 85.60 & 78.24 & 79.13 & 83.42 & 83.05 \\

\cmidrule(lr){3-16}

& & \multirow{2}{*}{384/16}
& \ding{55}
& 75.24 & 75.14 & 83.66 & 84.43 & 83.36 & 84.20 & 82.86 & 83.82 & 74.74 & 77.58 & 79.97 & 81.03 \\
& & & \ding{51}
& 83.74 & 82.85 & 90.02 & 89.88 & 88.86 & 88.79 & 89.34 & 89.24 & 84.60 & 85.18 & 87.31 & 87.19 \\

\cmidrule(lr){3-16}

& & \multirow{2}{*}{512/16}
& \ding{55}
& 76.58 & 76.90 & 83.54 & 84.88 & 83.74 & 85.02 & 83.79 & 85.11 & 77.30 & 80.24 & 80.99 & 82.43 \\
& & & \ding{51}
& 84.73 & 84.66 & 92.84 & 93.05 & 91.14 & 91.50 & 92.38 & 92.64 & 82.68 & 84.59 & 88.76 & 89.29 \\

\bottomrule
\end{tabular}
}

\caption{
Performance comparison of various CLIP-style vision foundation models on OpenSDI~\cite{wang2025opensdi}, evaluated with and without our tunable attention pooling (TAP). Best results are in \textbf{bold}, while second best are \underline{underlined}.
}
\label{tab:TAP_impact}
\end{table*}

In this section, We compare our approach to state-of-the-art methods on three challenging datasets: GenImage \cite{zhu2023genimage}, Chameleon \cite{yan2024sanity}, and OpenSDI \cite{wang2025opensdi}. GenImage and Chameleon consist of only fully-generated images, and share the same training data (162k images from SD v1.4 \cite{rombach2022ldm}+ 162k real images from ImageNet \cite{deng2009imagenet}). OpenSDI targets on detection of both fully-generated and inpainted images, where methods are trained on a mix of 100k real images from Megalith-10M \cite{bohan2024megalith10m} + 100k inpainted/fully-generated images from SD v1.5 \cite{rombach2022ldm}. In the Ablation Studies Section, we provide our comprehensive AIGI detection benchmark for comparing transferability of VFMs to this task. Moreover, we measure the impact of utilizing our TAP feature extraction method on the detection performance when combined with various ViT scales and input resolutions. 

\textbf{Implementation Details.}
Unless otherwise stated, we use either the ViT-L/14 or the ViT-G/14 pretrained models from Perception Encoder \cite{bolya2025perception} for semantic feature extraction. We refer the reader to Section \ref{ablations:vfm_benchmark} for more analysis regarding this choice. We use the AdamW optimizer with an iteration-based scheduler and set the learning rate at \(1e-4\), weight decay at \(1e-2\), beta range at \((0.9, 0.999)\), and use a batch size of 128. We train for a number of iterations equivalent to a single epoch on all training subsets, namely 2532 iterations for GenImage/Chameleon, and 1569 iterations for OpenSDI. For image augmentation, we use random JPEG compression with a quality range of (30,100) and a probability of 0.5, followed by random gaussian blur with a random kernel size of 3 or 5, and a probability of 0.5. For all our experiments, a single NVIDIA H100 GPU is used.

\subsection{State-of-the-Art Comparison}
\textbf{GenImage.} We compare our approach with and without TAP features against established methods on the respective test sets of GenImage. Table \ref{tab:main_results_genimage} shows the accuracy over the individual test sets, along with the mean accuracy. From the table, we observe that using the ViT-G/14 variant of Perception Encoder achieves a new state-of-the-art on the benchmark, outperforming prior best methods OMAT \cite{zhou2025breakingomat} and AIDE \cite{yan2024sanity} by 2.5\% and 10.2\% in accuracy, respectively. Evidently, we notice a sharp decline of generalization to Midjourney when employing TAP features. We hypothesize that the additional features may cause the final classifier to slightly overfit on artifacts more prevalent to SD v1.4 from the $gpl$ token, and lose track of the original $cls$ token. In this setting, shortening the number of iterations might help maintain the generalization from the frozen encoder.
\textbf{Chameleon.}
Table \ref{tab:main_results_chameleon} shows a comparison between our method and prior approaches on the challenging Chameleon dataset \cite{yan2024sanity} for in-the-wild AIGI detection. Our method easily achieves a new state-of-the-art performance, surpassing prior best method OMAT \cite{zhou2025breakingomat} by more than 29\% in detection accuracy. Here, utilizing TAP features improves accuracy by more than 10\% and 3\% when using PE-Core-ViT-L and PE-Core-ViT-G \cite{bolya2025perception}, respectively. 

\textbf{OpenSDI.} We compare our approach in detecting mixed inpainted/fully-generated images on the OpenSDI dataset \cite{wang2025opensdi} against prior methods in Table \ref{tab:main_results_opensdi}. Our approach once again achieves state-of-the-art performance, outperforming the previous best method, DualSight \cite{Abdullah2026DualSight}, by over 4\% in mean F1 score and 3\% in mean accuracy. Moreover, leveraging patch tokens through our TAP features proves advantageous for detecting inpainted images, yielding mean accuracy improvements of more than 5\% and 3\% compared to using only the $cls$ tokens of PE-ViT-L and PE-ViT-G, respectively.

\subsection{Ablation Studies}
\subsubsection{Benchmarking Vision Foundation Models}
\label{ablations:vfm_benchmark}
To obtain a clear picture about the transferability of vision foundation models to AIGI detection, we benchmark the following model families on OpenSDI \cite{wang2025opensdi} using only the frozen vision encoder, and a single linear classifier layer: DINOv2 \cite{oquab2023dinov2}, DINOv3 \cite{dinov3}, SAM2.1 \cite{sam2}, CLIP, \cite{radford2021clip}, SIGLIP2 \cite{siglip2}, and Perception Encoder (PE) \cite{bolya2025perception}. We pick OpenSDI as our benchmark dataset to quantify the sensitivity of the respective feature spaces to both fully-generated and -inpainted images. Each model family comes with a variety of options regarding model scale and input resolutions, which we take into consideration. Furthermore, model families such as DINOv2 and DINOv3 (as well as the NaFlex variant of SIGLIP2) have been trained to process variable-resolution images. This capability leads to the question of whether or not a higher resolution input image would provide a boost in detection performance, since more low-level semantic information is being extracted by virtue of a higher patch count. Since many AIGI detection methods that rely on CLIP utilize models at the scale of ViT-L or larger \cite{wang2025opensdi, koutlis2024rine, yan2024sanity, ojha2023unifd}, we restrict our benchmark to models of ViT-L scale or above.

In Table \ref{tab:visual_res}, we observe the following trends: (i) CLIP-style VFMs such as SIGLIP2 and PE outperform their self-supervised counterparts DINOv2 and DINOv3, while SAM2's feature space proves inadequate for AIGI detection, despite large-scale training. (ii) For fixed-resolution models, a higher resolution model typically outperforms its smaller resolution counterpart, given an equal model scale. (iii) For variable-resolution models, increasing the input image resolution rarely correlates with increased performance. On the contrary, we observe a decline in accuracy in the cases of DINOv2 and DINOv3 when scaling up input image sizes. Finally, out of all the VFMs we tested, the PE-Core models have recorded the highest F1 and accuracy scores in our benchmark, surpassing the standard CLIP by a wide margin. Our findings highlight a need for integrating a more capable model such as PE-Core as a standard for semantic feature extraction in the AIGI detection task.

\subsubsection{Impact of Tunable Attention-Pooling}
\label{TAP_impact}
In Table \ref{tab:TAP_impact}, we compare the detection performance of CLIP-style ViTs on OpenSDI \cite{wang2025opensdi} when combined with our tunable attention pooling (TAP). We observe significant performance gains across all variants and resolutions. Utilizing TAP scales well with high-resolution ViTs, as well as variable-resolution ViTs such as SIGLIP2-NaFlex. By exploiting the information within the patch tokens, we are able to extract fine-grained semantic patterns neglected by the $cls$ token, leading to improved accuracy.

\section{Conclusion}
\label{sec:conc} 
In this work, we systematically evaluate the transferability of modern vision foundation models (VFM) to the AIGI detection task, highlighting the need to adopt more recent VFMs as a standard for semantic feature extraction. Our benchmark demonstrates that newer models, particularly the Perception Encoder \cite{bolya2025perception}, provide substantially stronger out-of-the-box features for detecting both fully-generated and -inpainted images. Furthermore, we introduce tunable attention pooling (TAP), a lightweight semantic feature extraction mechanism that leverages the full token sequence of vision transformers rather than relying solely on the $cls$ token. By incorporating both patch tokens and the $cls$ token, TAP better aligns pretrained representations with the AIGI detection task while maintaining minimal trainable parameters. The resulting combination of modern VFMs and TAP features significantly improves generalization to unseen generators and establishes a new state of the art on challenging AIGI detection benchmarks.

{
    \small
    \bibliographystyle{ieeenat_fullname}
    \bibliography{main}

@String(IJCV = {Int. J. Comput. Vis.})

@String(CVPR= {IEEE Conf. Comput. Vis. Pattern Recog.})

@String(ICCV= {Int. Conf. Comput. Vis.})

@String(ECCV= {Eur. Conf. Comput. Vis.})

@String(ICPR = {Int. Conf. Pattern Recog.})

@String(ICLR = {Int. Conf. Learn. Represent.})

@String(AAAI = {AAAI})

@String(CVPRW= {IEEE Conf. Comput. Vis. Pattern Recog. Worksh.})

@String(IJCV  = {IJCV})

@String(CVPR  = {CVPR})

@String(ICCV  = {ICCV})

@String(ECCV  = {ECCV})

@String(ICPR  = {ICPR})

@String(ICLR  = {ICLR})

@String(CVPRW= {CVPRW})

@inproceedings{Abdullah2026DualSight,
    title = {DualSight: Learning to Disentangle Artifact and Semantic Features for Detection of Diffusion-Generated Images},
    author = {Abdullah, Ahmed and Ebert, Nikolas and Wasenm{\"u}ller, Oliver},
    booktitle = {International Conference on Pattern Recognition (ICPR)},
    year = {2026},
}

@inproceedings{wang2020cnn,
  title={CNN-generated images are surprisingly easy to spot... for now},
  author={Wang, Sheng-Yu and Wang, Oliver and Zhang, Richard and Owens, Andrew and Efros, Alexei A},
  booktitle={Conference on Computer Vision and Pattern Recognition (CVPR)},
  year={2020}
}

@inproceedings{liu2020gramnet,
  title={Global texture enhancement for fake face detection in the wild},
  author={Liu, Zhengzhe and Qi, Xiaojuan and Torr, Philip HS},
  booktitle={Conference on Computer Vision and Pattern Recognition (CVPR)},
  year={2020}
}

@inproceedings{tan2024freqnet,
  title={Frequency-aware deepfake detection: Improving generalizability through frequency space domain learning},
  author={Tan, Chuangchuang and Zhao, Yao and Wei, Shikui and Gu, Guanghua and Liu, Ping and Wei, Yunchao},
  booktitle={AAAI conference on Artificial Intelligence},
  year={2024}
}

@inproceedings{tan2024npr,
  title={Rethinking the up-sampling operations in cnn-based generative network for generalizable deepfake detection},
  author={Tan, Chuangchuang and Zhao, Yao and Wei, Shikui and Gu, Guanghua and Liu, Ping and Wei, Yunchao},
  booktitle={Conference on Computer Vision and Pattern Recognition (CVPR)},
  year={2024}
}

@inproceedings{ojha2023unifd,
  title={Towards universal fake image detectors that generalize across generative models},
  author={Ojha, Utkarsh and Li, Yuheng and Lee, Yong Jae},
  booktitle={Conference on Computer Vision and Pattern Recognition (CVPR)},
  year={2023}
}

@inproceedings{koutlis2024rine,
  title={Leveraging representations from intermediate encoder-blocks for synthetic image detection},
  author={Koutlis, Christos and Papadopoulos, Symeon},
  booktitle={European Conference on Computer Vision (ECCV)},
  year={2024},
}

@inproceedings{chen2021mvssnet,
  title={Image manipulation detection by multi-view multi-scale supervision},
  author={Chen, Xinru and Dong, Chengbo and Ji, Jiaqi and Cao, Juan and Li, Xirong},
  booktitle={International Conference on Computer Vision (ICCV)},
  year={2021}
}

@article{kwon2022catnet,
  title={Learning jpeg compression artifacts for image manipulation detection and localization},
  author={Kwon, Myung-Joon and Nam, Seung-Hun and Yu, In-Jae and Lee, Heung-Kyu and Kim, Changick},
  journal={International Journal of Computer Vision (IJCV)},
  year={2022},
}

@article{liu2022psccnet,
  title={PSCC-Net: Progressive spatio-channel correlation network for image manipulation detection and localization},
  author={Liu, Xiaohong and Liu, Yaojie and Chen, Jun and Liu, Xiaoming},
  journal={IEEE Transactions on Circuits and Systems for Video Technology},
  year={2022},
}

@inproceedings{wang2022objectformer,
  title={Objectformer for image manipulation detection and localization},
  author={Wang, Junke and Wu, Zuxuan and Chen, Jingjing and Han, Xintong and Shrivastava, Abhinav and Lim, Ser-Nam and Jiang, Yu-Gang},
  booktitle={Conference on Computer Vision and Pattern Recognition (CVPR)},
  year={2022}
}

@inproceedings{guillaro2023trufor,
  title={Trufor: Leveraging all-round clues for trustworthy image forgery detection and localization},
  author={Guillaro, Fabrizio and Cozzolino, Davide and Sud, Avneesh and Dufour, Nicholas and Verdoliva, Luisa},
  booktitle={Conference on Computer Vision and Pattern Recognition (CVPR)},
  year={2023}
}

@inproceedings{smeu2025declip,
  title={DeCLIP: Decoding CLIP representations for deepfake localization},
  author={Smeu, Stefan and Oneata, Elisabeta and Oneata, Dan},
  booktitle={Winter Conference on Applications of Computer Vision (WACV)},
  year={2025},
}

@article{ma2023imlvit,
  title={IML-ViT: Benchmarking image manipulation localization by vision transformer},
  author={Ma, Xiaochen and Du, Bo and Jiang, Zhuohang and Hammadi, Ahmed Y Al and Zhou, Jizhe},
  journal={arXiv preprint arXiv:2307.14863},
  year={2023}
}

@inproceedings{wang2025opensdi,
  title={OpenSDI: Spotting Diffusion-Generated Images in the Open World},
  author={Wang, Yabin and Huang, Zhiwu and Hong, Xiaopeng},
  booktitle={Conference on Computer Vision and Pattern Recognition (CVPR)},
  year={2025}
}

@inproceedings{radford2021clip,
  title={Learning transferable visual models from natural language supervision},
  author={Radford, Alec and Kim, Jong Wook and Hallacy, Chris and Ramesh, Aditya and Goh, Gabriel and Agarwal, Sandhini and Sastry, Girish and Askell, Amanda and Mishkin, Pamela and Clark, Jack and others},
  booktitle={International Conference on Machine Learning (ICML)},
  year={2021}
}

@inproceedings{zanella2024loravit,
  title={Low-Rank Few-Shot Adaptation of Vision-Language Models},
  author={Zanella, Maxime and Ben Ayed, Ismail},
  booktitle={Conference on Computer Vision and Pattern Recognition Workshops (CVPRW)},
  year={2024}
}

@article{ebert2023plg,
  title={{PLG-ViT}: Vision transformer with parallel local and global self-attention},
  author={Ebert, Nikolas and Stricker, Didier and Wasenm{\"u}ller, Oliver},
  journal={Sensors},
  volume={23},
  number={7},
  pages={3447},
  year={2023},
  publisher={MDPI}
}

@inproceedings{he2016resnet,
  title={Deep residual learning for image recognition},
  author={He, Kaiming and Zhang, Xiangyu and Ren, Shaoqing and Sun, Jian},
  booktitle={Conference on Computer Vision and Pattern Recognition (CVPR)},
  year={2016}
}

@inproceedings{liu2021swin,
  title={Swin transformer: Hierarchical vision transformer using shifted windows},
  author={Liu, Ze and Lin, Yutong and Cao, Yue and Hu, Han and Wei, Yixuan and Zhang, Zheng and Lin, Stephen and Guo, Baining},
  booktitle={Conference on Computer Vision and Pattern Recognition (CVPR)},
  year={2021}
}

@inproceedings{deng2009imagenet,
  title={Imagenet: A large-scale hierarchical image database},
  author={Deng, Jia and Dong, Wei and Socher, Richard and Li, Li-Jia and Li, Kai and Fei-Fei, Li},
  booktitle={Conference on Computer Vision and Pattern Recognition (CVPR)},
  year={2009}
}

@inproceedings{karras2019style,
  title={A style-based generator architecture for generative adversarial networks},
  author={Karras, Tero and Laine, Samuli and Aila, Timo},
  booktitle={Conference on Computer Vision and Pattern Recognition (CVPR)},
  year={2019}
}

@inproceedings{rombach2022ldm,
  title={High-resolution image synthesis with latent diffusion models},
  author={Rombach, Robin and Blattmann, Andreas and Lorenz, Dominik and Esser, Patrick and Ommer, Bj{\"o}rn},
  booktitle={Conference on Computer Vision and Pattern Recognition (CVPR)},
  year={2022}
}

@article{van2017vqvae,
  title={Neural discrete representation learning},
  author={Van Den Oord, Aaron and Vinyals, Oriol and others},
  journal={Neural Information Processing Systems (NeurIPS)},
  year={2017}
}

@article{razavi2019vqvae2,
  title={Generating diverse high-fidelity images with vq-vae-2},
  author={Razavi, Ali and Van den Oord, Aaron and Vinyals, Oriol},
  journal={Neural Information Processing Systems (NeurIPS)},
  year={2019}
}

@misc{flux2024,
    author={Black Forest Labs},
    title={FLUX},
    year={2024},
    howpublished={\url{https://github.com/black-forest-labs/flux}},
}

@inproceedings{zhang2023addingcontrolnet,
  title={Adding conditional control to text-to-image diffusion models},
  author={Zhang, Lvmin and Rao, Anyi and Agrawala, Maneesh},
  booktitle={International Conference on Computer Vision (ICCV)},
  year={2023}
}

@inproceedings{tan2025c2pclip,
  title={C2p-clip: Injecting category common prompt in clip to enhance generalization in deepfake detection},
  author={Tan, Chuangchuang and Tao, Renshuai and Liu, Huan and Gu, Guanghua and Wu, Baoyuan and Zhao, Yao and Wei, Yunchao},
  booktitle={AAAI conference on Artificial Intelligence},
  year={2025}
}

@inproceedings{karras2018progressiveprogan,
  title={Progressive Growing of GANs for Improved Quality, Stability, and Variation},
  author={Karras, Tero and Aila, Timo and Laine, Samuli and Lehtinen, Jaakko},
  booktitle={International Conference on Learning Representations (ICLR)},
  year={2018}
}

@inproceedings{dosovitskiy2021vit,
  title={An Image is Worth 16x16 Words: Transformers for Image Recognition at Scale},
  author={Dosovitskiy, Alexey and Beyer, Lucas and Kolesnikov, Alexander and Weissenborn, Dirk and Zhai, Xiaohua and Unterthiner, Thomas and Dehghani, Mostafa and Minderer, Matthias and Heigold, Georg and Gelly, Sylvain and others},
  booktitle={International Conference on Learning Representations (ICLR)},
  year={2021}
}

@inproceedings{hulorallm,
  title={LoRA: Low-Rank Adaptation of Large Language Models},
  author={Hu, Edward J and Wallis, Phillip and Allen-Zhu, Zeyuan and Li, Yuanzhi and Wang, Shean and Wang, Lu and Chen, Weizhu and others},
  booktitle={International Conference on Learning Representations (ICLR)},
  year={2022}
}

@inproceedings{wang2023dire,
  title={Dire for diffusion-generated image detection},
  author={Wang, Zhendong and Bao, Jianmin and Zhou, Wengang and Wang, Weilun and Hu, Hezhen and Chen, Hong and Li, Houqiang},
  booktitle={International Conference on Computer Vision (ICCV},
  year={2023}
}

@inproceedings{zhang2019detecting,
  title={Detecting and simulating artifacts in gan fake images},
  author={Zhang, Xu and Karaman, Svebor and Chang, Shih-Fu},
  booktitle={IEEE international workshop on information forensics and security (WIFS)},
  year={2019},
}

@inproceedings{frank2020leveragingfreq,
  title={Leveraging frequency analysis for deep fake image recognition},
  author={Frank, Joel and Eisenhofer, Thorsten and Sch{\"o}nherr, Lea and Fischer, Asja and Kolossa, Dorothea and Holz, Thorsten},
  booktitle={International Conference on Machine Learning (ICML)},
  year={2020},
}

@inproceedings{touvron2021deit,
  title={Training data-efficient image transformers \& distillation through attention},
  author={Touvron, Hugo and Cord, Matthieu and Douze, Matthijs and Massa, Francisco and Sablayrolles, Alexandre and J{\'e}gou, Herv{\'e}},
  booktitle={International Conference on Machine Learning (ICML)},
  year={2021}
}

@inproceedings{zhang2019detectingspec,
  title={Detecting and simulating artifacts in gan fake images},
  author={Zhang, Xu and Karaman, Svebor and Chang, Shih-Fu},
  booktitle={International Workshop on Information Forensics and Security (WIFS)},
  year={2019},
  organization={IEEE}
}

@inproceedings{qian2020thinkingf3net,
  title={Thinking in frequency: Face forgery detection by mining frequency-aware clues},
  author={Qian, Yuyang and Yin, Guojun and Sheng, Lu and Chen, Zixuan and Shao, Jing},
  booktitle={European Conference on Computer Vision (ECCV)},
  year={2020},
  organization={Springer}
}

@inproceedings{chollet2017xception,
  title={Xception: Deep learning with depthwise separable convolutions},
  author={Chollet, Fran{\c{c}}ois},
  booktitle={Conference on Computer Vision and Pattern Recognition (CVPR)},
  year={2017}
}

@article{zhu2023gendet,
  title={Gendet: Towards good generalizations for ai-generated image detection},
  author={Zhu, Mingjian and Chen, Hanting and Huang, Mouxiao and Li, Wei and Hu, Hailin and Hu, Jie and Wang, Yunhe},
  journal={arXiv preprint arXiv:2312.08880},
  year={2023}
}

@article{zhong2023patchcraft,
  title={Patchcraft: Exploring texture patch for efficient ai-generated image detection},
  author={Zhong, Nan and Xu, Yiran and Li, Sheng and Qian, Zhenxing and Zhang, Xinpeng},
  journal={arXiv preprint arXiv:2311.12397},
  year={2023}
}

@inproceedings{liu2021spatial,
  title={Spatial-phase shallow learning: rethinking face forgery detection in frequency domain},
  author={Liu, Honggu and Li, Xiaodan and Zhou, Wenbo and Chen, Yuefeng and He, Yuan and Xue, Hui and Zhang, Weiming and Yu, Nenghai},
  booktitle={Conference on Computer Vision and Pattern Recognition (CVPR)},
  year={2021}
}

@inproceedings{luo2021generalizingsrm,
  title={Generalizing face forgery detection with high-frequency features},
  author={Luo, Yuchen and Zhang, Yong and Yan, Junchi and Liu, Wei},
  booktitle={Conference on Computer Vision and Pattern Recognition (CVPR)},
  pages={16317--16326},
  year={2021}
}

@inproceedings{zhou2025breakingomat,
  title={Breaking latent prior bias in detectors for generalizable aigc image detection},
  author={Zhou, Yue and He, Xinan and Lin, KaiQing and Fan, Bin and Ding, Feng and Li, Bin},
  booktitle={Neural Information Processing Systems (NeurIPS)},
  year={2025}
}

@inproceedings{yan2024sanity,
  title={A sanity check for ai-generated image detection},
  author={Yan, Shilin and Li, Ouxiang and Cai, Jiayin and Hao, Yanbin and Jiang, Xiaolong and Hu, Yao and Xie, Weidi},
  booktitle={International Conference on Learning Representations (ICLR)},
  year={2025}
}

@inproceedings{bolya2025perception,
  title={Perception encoder: The best visual embeddings are not at the output of the network},
  author={Bolya, Daniel and Huang, Po-Yao and Sun, Peize and Cho, Jang Hyun and Madotto, Andrea and Wei, Chen and Ma, Tengyu and Zhi, Jiale and Rajasegaran, Jathushan and Rasheed, Hanoona and others},
  booktitle={Neural Information Processing Systems (NeurIPS)},
  year={2025}
}

@inproceedings{liu2022detectinglnp,
  title={Detecting generated images by real images},
  author={Liu, Bo and Yang, Fan and Bi, Xiuli and Xiao, Bin and Li, Weisheng and Gao, Xinbo},
  booktitle={European Conference on Computer Vision (ECCV)},
  year={2022},
  organization={Springer}
}

@article{siglip2,
  title={Siglip 2: Multilingual vision-language encoders with improved semantic understanding, localization, and dense features},
  author={Tschannen, Michael and Gritsenko, Alexey and Wang, Xiao and Naeem, Muhammad Ferjad and Alabdulmohsin, Ibrahim and Parthasarathy, Nikhil and Evans, Talfan and Beyer, Lucas and Xia, Ye and Mustafa, Basil and others},
  journal={arXiv preprint arXiv:2502.14786},
  year={2025}
}

@inproceedings{zhu2023genimage,
  title={Genimage: A million-scale benchmark for detecting ai-generated image},
  author={Zhu, Mingjian and Chen, Hanting and Yan, Qiangyu and Huang, Xudong and Lin, Guanyu and Li, Wei and Tu, Zhijun and Hu, Hailin and Hu, Jie and Wang, Yunhe},
  booktitle={Neural Information Processing Systems (NeurIPS)},
  year={2023}
}

@article{oquab2023dinov2,
  title={Dinov2: Learning robust visual features without supervision},
  author={Oquab, Maxime and Darcet, Timoth{\'e}e and Moutakanni, Th{\'e}o and Vo, Huy and Szafraniec, Marc and Khalidov, Vasil and Fernandez, Pierre and Haziza, Daniel and Massa, Francisco and El-Nouby, Alaaeldin and others},
  journal={arXiv preprint arXiv:2304.07193},
  year={2023}
}

@inproceedings{he2022masked,
  title={Masked autoencoders are scalable vision learners},
  author={He, Kaiming and Chen, Xinlei and Xie, Saining and Li, Yanghao and Doll{\'a}r, Piotr and Girshick, Ross},
  booktitle={Conference on Computer Vision and Pattern Recognition (CVPR)},
  year={2022}
}

@article{su2024roformer,
  title={Roformer: Enhanced transformer with rotary position embedding},
  author={Su, Jianlin and Ahmed, Murtadha and Lu, Yu and Pan, Shengfeng and Bo, Wen and Liu, Yunfeng},
  journal={Neurocomputing},
  volume={568},
  pages={127063},
  year={2024},
  publisher={Elsevier}
}

@inproceedings{beyer2023flexivit,
  title={Flexivit: One model for all patch sizes},
  author={Beyer, Lucas and Izmailov, Pavel and Kolesnikov, Alexander and Caron, Mathilde and Kornblith, Simon and Zhai, Xiaohua and Minderer, Matthias and Tschannen, Michael and Alabdulmohsin, Ibrahim and Pavetic, Filip},
  booktitle={Conference on Computer Vision and Pattern Recognition (CVPR)},
  year={2023}
}

@inproceedings{dehghani2023patch,
  title={Patch n’pack: Navit, a vision transformer for any aspect ratio and resolution},
  author={Dehghani, Mostafa and Mustafa, Basil and Djolonga, Josip and Heek, Jonathan and Minderer, Matthias and Caron, Mathilde and Steiner, Andreas and Puigcerver, Joan and Geirhos, Robert and Alabdulmohsin, Ibrahim M and others},
  booktitle={Neural Information Processing Systems (NeurIPS)},
  year={2023}
}

@article{dinov3,
  title={Dinov3},
  author={Sim{\'e}oni, Oriane and Vo, Huy V and Seitzer, Maximilian and Baldassarre, Federico and Oquab, Maxime and Jose, Cijo and Khalidov, Vasil and Szafraniec, Marc and Yi, Seungeun and Ramamonjisoa, Micha{\"e}l and others},
  journal={arXiv preprint arXiv:2508.10104},
  year={2025}
}

@inproceedings{sam2,
  title={SAM 2: Segment Anything in Images and Videos},
  author={Ravi, Nikhila and Gabeur, Valentin and Hu, Yuan-Ting and Hu, Ronghang and Ryali, Chaitanya and Ma, Tengyu and Khedr, Haitham and R{\"a}dle, Roman and Rolland, Chloe and Gustafson, Laura and others},
  booktitle={International Conference on Learning Representations (ICLR)},
  year={2025}
}

@inproceedings{peebles2023scalabledit,
  title={Scalable diffusion models with transformers},
  author={Peebles, William and Xie, Saining},
  booktitle={International Conference on Computer Vision (ICCV)},
  year={2023}
}

@inproceedings{zhai2022scaling,
  title={Scaling vision transformers},
  author={Zhai, Xiaohua and Kolesnikov, Alexander and Houlsby, Neil and Beyer, Lucas},
  booktitle={Conference on Computer Vision and Pattern Recognition (CVPR)},
  year={2022}
}

@misc{bohan2024megalith10m,
  author       = {Ollin Boer Bohan},
  title        = {Megalith-10m: A dataset of public domain photographs},
  year         = {2024},
  howpublished = {\url{https://huggingface.co/datasets/madebyollin/megalith-10m}},
  note         = {Accessed: 2026-02-26}
}

@misc{flux-2,
    author={Black Forest Labs},
    title={{FLUX.2: Frontier Visual Intelligence}},
    year={2025},
    howpublished={\url{https://bfl.ai/blog/flux-2}},
}

@inproceedings{yermakov2026deepfakeclip,
  title={Deepfake detection that generalizes across benchmarks},
  author={Yermakov, Andrii and Cech, Jan and Matas, Jiri and Fritz, Mario},
  booktitle={Winter Conference on Applications of Computer Vision (WACV)},
  year={2026}
}

@inproceedings{vaswani2017attention,
  title={Attention is all you need},
  author={Vaswani, Ashish and Shazeer, Noam and Parmar, Niki and Uszkoreit, Jakob and Jones, Llion and Gomez, Aidan N and Kaiser, {\L}ukasz and Polosukhin, Illia},
  booktitle={Neural Information Processing Systems (NeurIPS)},
  year={2017}
}

@misc{midjourney,
  title        = {Midjourney},
  author       = {{Midjourney, Inc.}},
  year         = {2022},
  howpublished = {\url{https://www.midjourney.com/home/}},
}

@article{zhou2025broughtgun,
  title={Brought a gun to a knife fight: Modern vfm baselines outgun specialized detectors on in-the-wild ai image detection},
  author={Zhou, Yue and He, Xinan and Lin, Kaiqing and Fan, Bing and Ding, Feng and Zeng, Jinhua and Li, Bin},
  journal={arXiv preprint arXiv:2509.12995},
  year={2025}
}

@inproceedings{podell2023sdxl,
  title={Sdxl: Improving latent diffusion models for high-resolution image synthesis},
  author={Podell, Dustin and English, Zion and Lacey, Kyle and Blattmann, Andreas and Dockhorn, Tim and M{\"u}ller, Jonas and Penna, Joe and Rombach, Robin},
  booktitle={International Conference on Learning Representations (ICLR)},
  year={2024}
}

@inproceedings{esser2024scalingsd3,
  title={Scaling rectified flow transformers for high-resolution image synthesis},
  author={Esser, Patrick and Kulal, Sumith and Blattmann, Andreas and Entezari, Rahim and M{\"u}ller, Jonas and Saini, Harry and Levi, Yam and Lorenz, Dominik and Sauer, Axel and Boesel, Frederic and others},
  booktitle={International Conference on Machine Learning (ICML)},
  year={2024}
}
}


\end{document}